\documentclass[preprint,11pt,authoryear]{elsarticle}

\usepackage[T1]{fontenc}
\usepackage{graphicx}
\usepackage{float}
\usepackage{booktabs}
\usepackage{placeins}
\usepackage{caption} 
\usepackage{subcaption}
\usepackage{multirow}

\usepackage[a4paper,margin=1in]{geometry}
\usepackage{xcolor}
\usepackage{comment}
\excludecomment{hidden}

\usepackage{algorithm}
\usepackage{algpseudocode}

\usepackage{amsmath, amssymb}

\definecolor{accent}{RGB}{0, 114, 189} 
\definecolor{dark}{RGB}{33, 37, 41}    

\usepackage{float}

\usepackage[linesnumbered, ruled, vlined, algo2e]{algorithm2e}

\definecolor{colorCLK}{RGB}{31,119,180}
\definecolor{colorEAX}{RGB}{255,127,14}
\definecolor{colorLKH}{RGB}{44,160,44}
\definecolor{colorMAOS}{RGB}{214,39,40}
\definecolor{colorCONCORDE}{RGB}{148,103,189}

\usepackage{pgfplots}
\usepgfplotslibrary{groupplots}

\algnewcommand\Input{\State \textbf{Input: }}
\algnewcommand\Output{\State \textbf{Output: }}

\usepackage{amsmath, amssymb}

\usepackage{hyperref}

\usepackage{enumitem}

\usepackage{listings}

\usepackage{tikz}

\usepackage{pgfplots}
\pgfplotsset{compat=1.18}
\usetikzlibrary{patterns} 

\usepgfplotslibrary{statistics}
\usepackage{subcaption}

\usepackage[table]{xcolor}

\usepackage{pgfplotstable}

\usepackage{rotating}

\begin{document}

\begin{frontmatter}

%
\title{Graph Neural Network-based Algorithm Selection for the Traveling Salesman Problem: A Systematic Study of Cost and Rank Losses under Distinct Budget Regimes}
%
%

\author{Zhaoxuan Li\fnref{equal}}
\ead{zhaoxuan.li@dukekunshan.edu.cn}

\author{Jiale Yang\fnref{equal}}
\ead{jiale.yang@dukekunshan.edu.cn}

\author{Yifei Lu}
\ead{y.lu@dukekunshan.edu.cn}

\author{Mustafa M\i s\i r\corref{cor1}}
\ead{mustafa.misir@dukekunshan.edu.cn}

\fntext[equal]{These authors contributed equally to this work.}

\cortext[cor1]{Corresponding author}

\address{Duke Kunshan University, 8 Duke Avenue, Kunshan, 215316, China}



\begin{abstract}

Automated Algorithm Selection (AS) aims to improve problem-solving performance by selecting, for each problem instance, the most suitable algorithm from a predefined portfolio. It is particularly relevant to the Traveling Salesman Problem (TSP), where solver performance is strongly instance-dependent. We introduce GNNAS-TSP, a Graph Neural Network (GNN)-based AS framework that learns TSP instance representations directly from raw graph data, avoiding manual feature engineering. GNNAS-TSP formulates AS as a joint cost-prediction and ranking task. We evaluate cost-based (mean squared error (MSE), mean absolute error (MAE), and Huber), rank-based (RankNet, ListNet, and LambdaRank), and hybrid learning objectives for a portfolio comprising Chained Lin--Kernighan, Edge Assembly Crossover, Lin--Kernighan--Helsgaun, Multiagent Optimization System, and Concorde. Experiments use fixed computational budgets of 10 and 60 seconds. On the held-out test set, the selected configurations improve on the Single Best Solver (SBS) in normalized solution cost at both budgets.
For the 10s budget, AS achieves substantial and statistically significant cost improvement over SBS. 
Overall, the results suggest that GNNAS-TSP is a useful meta-solving strategy when exploitable variation exists across solver performance.

\end{abstract}


\end{frontmatter}

\section{Introduction}
\label{sec:intro}

Algorithm design is a fundamental task across disciplines, extending beyond Computer Science. 
However, both theoretical \citep{wolpert1997nofreelunch} and empirical studies highlight that no universal algorithm can consistently solve diverse problem instances effectively across extensive scenarios. 
This raises critical questions about the necessity of investing financial, human, and computational resources into designing new algorithms. 
Automated Algorithm Selection (AS) \citep{kerschke2019automated} is an alternative approach leveraging the strengths of multiple candidate algorithms instead of relying on a single one.
The focus of AS is on the optimized utilization of existing algorithms rather than the design of new ones.
The core idea is to automatically match the most suitable algorithm to each problem instance. 
Traditional AS achieves this by employing performance prediction models that estimate the performance of the present candidate algorithms on a given problem instance. 
The algorithm predicted to perform best is then chosen and applied as the solver.
This process effectively frames the AS task as a recommendation problem \citep{misir2017alors} where the objective is to suggest the most suitable algorithm for a given instance.

Routing is among the most extensively studied optimization problems in both academic and practical contexts. 
This has led to the development of numerous problem variants, publicly available instances and a wide range of algorithms. 
AS has also been applied to routing problems \citep{wagner2018case,misir2022algorithm,huerta2022aaas,heins2023study}. 
However, existing AS approaches for routing typically rely on explicitly engineered features to represent problem instances, which are then used to build AS models. 
Beyond AS, TSP has been targeted by a range of algorithmic approaches.
Like in many other fields, the present trend is the utilization of Machine Learning (ML) \citep{alanzi2025solving}. 
Within this domain, Graph Neural Networks (GNNs) \citep{song2025revisit,song2025geometrically} are particularly well-suited, as they operate directly on the graph representations of the TSP instances.
Building upon the suitability of GNNs for learning structural representations of combinatorial optimization problems, and the success of GNN-based AS systems on graph-structured tasks such as molecular docking \citep{ym2024gnnasdock,cao2025mcgnnasdock}, this work proposes a GNN-based AS approach for the Traveling Salesman Problem (TSP) called \textsc{GNNAS-TSP}. 

That being said, there is an earlier work using GNN-based AS for the TSP \citep{song2025revisit}. They proposed GINES, a GNN model modified based on Graph Isomorphism Network with Edge features, which formulates algorithm selection as a binary graph-classification task. In contrast, our framework predicts the performance of each candidate solver and selects the solver with the lowest predicted cost. This regression-based formulation accommodates solver portfolios with more than two candidates and permits learning objectives based on both predicted cost and relative rank. Because the two studies use different solver portfolios, performance targets, and evaluation protocols, their reported numerical results are not directly comparable. We therefore position GINES as a closely related methodological reference rather than as an empirical baseline.





The main contributions of this work are threefold: 
$(i)$ we propose a GNN-based AS framework for the TSP that learns instance representations directly from raw graph data; 
$(ii)$ we systematically study the impact of cost-based and rank-based loss functions, including their combinations; and 
$(iii)$ we provide an extensive empirical analysis of solver selection behavior and low-dimensional performance structure under different computational budgets.
The reported results on 300 TSP test instances show that GNNAS-TSP delivers performance improvements over the use of the constituent TSP solvers under both runtime budgets.

In the remainder of the paper, Section~\ref{sec:background} discusses both about AS and TSP referring to the existing literature.
Section~\ref{sec:methodology} introduces the GNNAS-TSP architecture and setup.
Section~\ref{sec:results} reports the experimental results while offering an in-depth analysis both on algorithms and instances.
Section~\ref{sec:conclusiont_futurework} delivers an overall summary, reports the key findings, and outlines ongoing as well as future research directions.

\begin{hidden}

\section{Background}
\label{sec:background}

\subsection{Algorithm Selection}
\label{sec:algosel}

Algorithm Selection (AS) \citep{rice1976algorithm,kerschke2019automated} addresses the fact that algorithm performance is often highly
instance-dependent: different solvers can excel on different regions of the instance space, so no
single algorithm is uniformly best across all instances. Formally, let $\mathcal{I}$ be a finite set of
problem instances, $\mathcal{P}=\{A_1,\dots,A_m\}$ a portfolio of candidate algorithms, and
$\rho:\mathcal{P}\times\mathcal{I}\rightarrow\mathbb{R}$ a performance measure (e.g., runtime, error, or
solution quality). Let $f:\mathcal{I}\rightarrow \mathbb{R}^d$ denote an instance representation (often
hand-crafted features). The AS task is to learn a selector $S:\mathbb{R}^d\rightarrow \mathcal{P}$ that
minimizes expected performance over an instance distribution $\mathcal{D}$:
\[
\mathbb{E}_{i\sim\mathcal{D}}\left[\rho\!\left(S(f(i)),\, i\right)\right].
\]
The virtual best solver (VBS) or oracle,
$S_{\text{VBS}}(i)=\arg\min_{A\in\mathcal{P}}\rho(A,i)$, provides a per-instance lower bound on
achievable performance while the single best solver (SBS) is the best fixed algorithm in
$\mathcal{P}$ when used on all instances. 
In other words, SBS refers to the traditional algorithm efforts of using one specific algorithm for solving all the target problem instances.
AS aims to narrow the gap between SBS and VBS by learning to exploit regularities in how instance properties relate to solver performance. 
A typical AS workflow is illustrated in Figure~\ref{fig:alg-selection}.

\begin{figure}[!ht]
\centering
\includegraphics[width=1\textwidth]{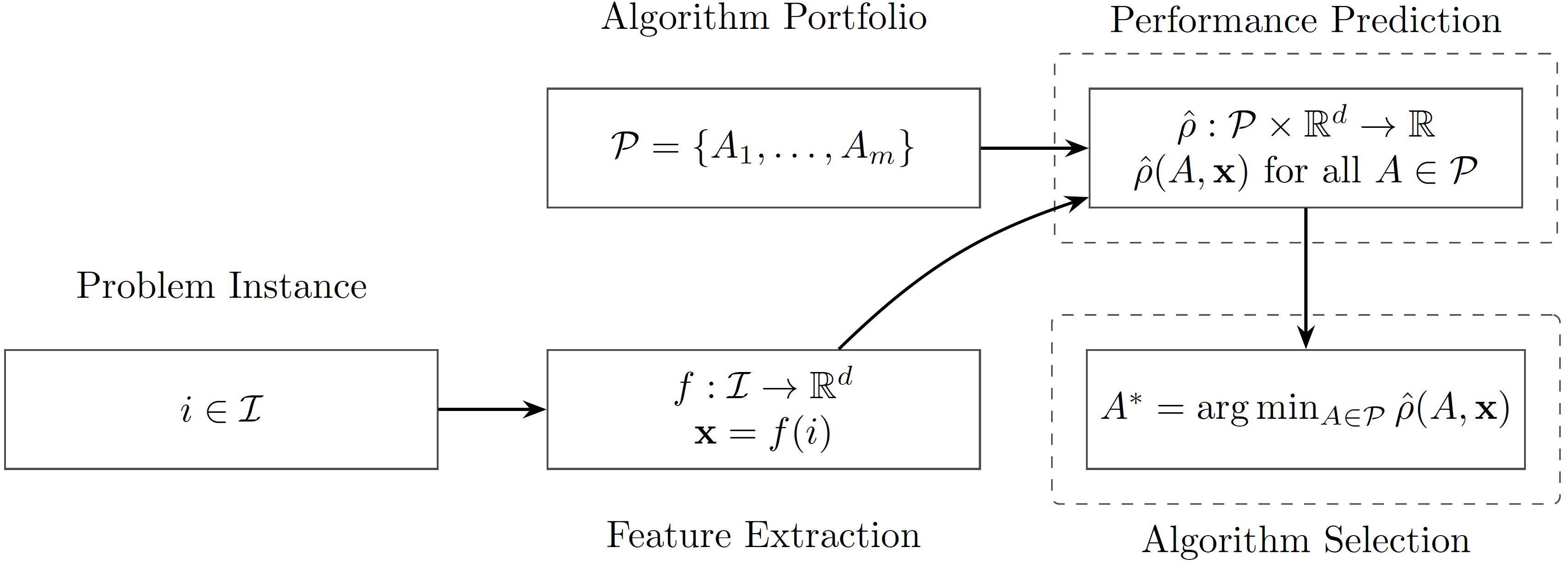}
\caption{A standard Algorithm Selection (AS) pipeline: represent the instance, predict relative solver
performance, and select the most promising algorithm.}
\label{fig:alg-selection}
\end{figure}

AS is rarely used for precise performance estimation yet identifying a near-best solver. 
This distinction motivates ranking-oriented training objectives and evaluation criteria beyond raw prediction
error especially when multiple solvers have very similar outcomes on many instances.
In this sense,

\paragraph{Benchmarks and the feature engineering bottleneck}
AS research has been strongly shaped by benchmark suites such as ASlib \citep{bischl2016aslib}, which
provide standardized scenarios with fixed portfolios, performance data, and predefined instance features.
While this has enabled systematic comparison of selection methods, it also highlights a key practical
bottleneck: high-quality hand-crafted features can be expensive to design and compute, and may not
transfer robustly across domains or instance distributions \citep{kerschke2019automated}.

To situate our approach, we briefly outline three representative AS paradigms that differ in how they represent instances and exploit historical performance data.

\paragraph{General (feature-based) AS}
Classical AS follows Rice's formulation \citep{rice1976algorithm}: an instance is mapped to an explicit
feature vector $f(i)$ and a model predicts which algorithm in a portfolio will perform best.
Most systems adopt supervised learning using historical performance matrices, and implement selection
via (i) per-algorithm performance prediction like a regression task of runtime or cost prediction, followed by choosing the predicted best solver, (ii) direct multi-class classification where the label is the oracle-best solver,
or (iii) learning-to-rank models that optimize relative solver ordering rather than absolute values
\citep{Kotthoff2014AlgorithmSelection,xu2008satzilla}. Benchmark suites such as ASlib standardize
evaluation by providing portfolios, performance data, and predefined features \citep{bischl2016aslib}.
While effective, feature-based pipelines often depend on domain expertise and can incur non-trivial
feature computation costs, limiting transfer across domains and instance distributions
\citep{kerschke2019automated}. These limitations motivate approaches that reduce reliance on manual
features or leverage performance structure more directly.

\paragraph{GNN-based AS}
Representation learning addresses the feature-engineering bottleneck by learning instance embeddings
from raw problem structure. For graph-structured domains, GNNs provide a strong inductive bias by
encoding instances as graphs and learning embeddings through message passing that aggregates local
and global structural information \citep{scarselli2008graph,wu2020comprehensive}. In GNN-based AS,
the learned graph embedding is mapped to predicted per-solver performance values or ranking scores,
enabling end-to-end training without hand-crafted instance features. This is particularly attractive in
combinatorial optimization, where structure is explicit (variables/constraints, cities/edges) but
effective handcrafted descriptors are often costly or incomplete. Recent studies demonstrate that
GNN-based selectors can generalize across instance distributions and can be extended to settings such as
budget-aware or multi-criteria selection \citep{ym2024gnnasdock,cao2025mcgnnasdock}. However, their
effectiveness depends on modeling choices (graph construction, pooling/readout) and, critically, the
learning objective: selection quality is governed by relative solver ordering, especially when solvers
are close in absolute performance.

\paragraph{ALORS (latent-factor AS)}
Latent-factor approaches view AS through the lens of recommender systems and focus on the
instance--algorithm performance matrix. ALORS factorizes this matrix into low-dimensional latent
representations of instances and algorithms, capturing implicit solver correlations and shared
performance structure \citep{misir2017alors}. This perspective can be effective even when instance
descriptors are weak or expensive, because the model exploits regularities in observed solver behavior
(e.g., groups of instances for which the same subset of solvers tends to be strong). For unseen instances,
ALORS projects available descriptors into the latent space to enable performance prediction and
recommendation. Compared with feature-based AS, latent-factor models emphasize the geometry of the
performance landscape; compared with graph-based representation learning, they rely less on explicit
instance structure and more on historical performance data. Conceptually, ALORS motivates analyzing
whether solver outcomes admit a low-dimensional structure---an idea that aligns with performance-space
analyses (e.g., SVD of rank profiles) and complements structure-aware encoders such as GNNs.

\paragraph{Beyond hand-crafted features: learned representations and latent performance structure}
To reduce reliance on manual feature design, recent work explores representation learning for AS, including CNN-based pipelines that learn from raw instance encodings \citep{loreggia2016deep}, and more recent embedding-based approaches (e.g., Transformer-derived representations) \citep{wang2025molecular}. For graph-structured problems, Graph Neural Networks (GNNs) offer a natural inductive bias by learning instance embeddings through message passing on nodes and edges \citep{scarselli2008graph,wu2020comprehensive}. Separately, latent-factor approaches such as ALORS model the instance--algorithm performance matrix directly via matrix factorization, capturing implicit structure in historical performance data without requiring a rich hand-crafted feature set \citep{misir2017alors}. These directions are complementary:
representation learning leverages explicit instance structure, while latent-factor models exploit solver correlations and low-dimensional structure in performance landscapes.

\paragraph{Relevance to our setting}
The TSP provides a natural testbed for AS because high-performing solvers exhibit substantial instance-wise variability, and their relative ordering can change with the available runtime budget.
In our benchmark, the identity of the top best solvers is not fixed across instances nor stable across budgets, which makes per-instance selection meaningful. Consequently, effective selectors should (i) learn instance representations that capture structural differences among TSP graphs, and (ii) optimize objectives that reflect selection quality—i.e., getting the solver ordering right—rather than only minimizing numerical prediction error. These observations motivate our GNN-based, ranking-oriented AS framework and our analysis of how solver performance structure differs between the 10s and 60s settings.

To the best of our knowledge, there are three existing works that use deep learning methods on TSP AS.
\citep{huerta2022aaas} uses CNN to generate an image representing the locations of the cities and uses the image as the model input. The model's output is Staggered Representation, a representation like ranking calculated based on algorithm's costs at different time steps. This representation ensures that every two algorithms have different values, and when two algorithms achieve the same result at a given time, the one that achieves it earilier as a higher "score". This also avoids the influence of different scale of costs due to different graph size. However, an intermediate image still needed to be generated, which makes the selection process not so convenient and can introduce irrelevant information. In comparison, TSP naturally has a graph structure and can be easily represented as a graph and be learned through GNN.

\citep{song2025ginestspas}
GINES extends the expressive GINE architecture by incorporating spatial distribution information, inspired by GNNs for point cloud processing. Specifically, node features are enriched with relative positional differences between neighboring cities, i.e., $\mathbf{p}_j - \mathbf{p}_i$, enabling the model to capture local geometric structures.

The model takes city coordinates and distances as input, where only the 10 nearest neighbors of each node are considered, and automatically extracts features without the need to handcraft features or generate intermediate representations. It employs three GNN layers, each followed by graph-level pooling, which capture different levels of spatial information, from local to global. The pooled representations are then concatenated to retain information at different scales.

 On the TSP-ISA dataset (1330 instances with 100 cities), GINES outperforms Random Forest (RF) and Graph Convolutional Networks (GCN); while on the TSP-CNN dataset (1000 instances with 1000 cities), it slightly outperforms CNN-based methods but underperforms RF, suggesting that despite its ability to learn directly from the raw graph representations, GINES still fails to extract some important features compared with RF using elaborately designed features. 

\citep{song2026iegnntspas}

Both these studies formulate the AS problem as a graph-level classification problem, and are evaluated on two datasets each involving selection between two candidate algorithms. Performance is measured using accuracy, defined as the frequency with which the model correctly selects the best-performing algorithm. When the portfolio size gets bigger, classification might not work so well.
(\citep{song2026iegnntspas} is also evaluated on hardness prediction)

\subsection{Traveling Salesman Problem (TSP)} 
\label{sec:tsp}


The Traveling Salesman Problem (TSP) is a classic NP-hard combinatorial optimization problem with fundamental significance across multiple academic and industrial domains such as operations research, computer science, and artificial intelligence.
TSP aims to find the minimum-cost Hamiltonian cycle in a complete edge-weighted graph $G = (V, E)$, where $V$ represents $n$ vertices (cities) and $E$ contains edges (roads) between all vertex pairs with associated costs.
The objective is to visit each city exactly once and return to the starting city while minimizing the total travel cost.
Due to its NP-hard nature, no polynomial-time algorithm exists that guarantees optimal solutions for general TSP instances.
Consequently, various approximation algorithms and heuristic methods have been developed that efficiently produce high-quality solutions, though not necessarily optimal, for practical applications in these diverse domains.

\textcolor{red}{TBA: algorithms}

\textcolor{red}{TBA: bechmarks--TSPLIB etc.}

\textcolor{red}{
TBA: Revise --
Based on the mathematical foundations, numerous heuristics have been developed to solve the Traveling Salesman Problem (TSP). 
However, the performance of these heuristics can vary significantly depending on factors such as population size and the structure of problem instances. 
Therefore, to comprehensively evaluate the effectiveness of a TSP heuristic, it is essential to test it on a diverse set of instances that exhibit a wide range of characteristics \citep{baniasadi2018new}. A widely used benchmark for this purpose is TSPLIB \citep{reinelt1991tsplib}, which was introduced in 1991. It provides a collection of TSP instances originating from various sources and encompassing different types \citep{reinhelt2014tsplib}]. This benchmark set is commonly used by researchers to compare the performance of new algorithms against established solutions.

TSP has likewise been a subject of AS research, reflecting its central importance in the field \citep{kotthoff2015improving}.

Since there are lots of benchmark sets for TSP, such as TSPLIB, numerous improved heuristics and algorithms have been developed for solving TSP near optimally. \citep{wang2007improved} introduced untwist operator to \textit{Genetic Algorithm} to untie the knots of route effectively, shortening the length of route and boosting the convergent rate. \citep{sanches2017improving} proposed the The GPX2 partition crossover operator to improve the initial solution for the \textit{Concorde branch-and-cut algorithm} by efficiently combining multiple local optima found through iterated local search. \citep{d2020learning} introduced a reinforcement learning-based approach for solving the Euclidean TSP, which learns a stochastic policy by using historical search information. The method achieves near-optimal solutions, outperforms previous deep learning and heuristic approaches. All these methods require the artificial guidance in order to find the expected tour efficiently. Additionally, the design of heuristics requires researchers have substantial domain-specific knowledge. The first giant leap for tackling these challenges is under the development of \textit{Deep Learning}. Deep learning can automatically assign different heuristics to the different instances based on the task, thus reducing human-need for heuristic design \citep{bello2016neural}, \citep{kool2018attention}.

TBA: 
\begin{itemize}
    \item formal TSP description (+ NP-hard etc.)
    \item existing algorithms ++ GNN based ones
    \item variants
\end{itemize}


The \emph{Traveling Salesman Problem} (TSP) is a canonical NP-hard combinatorial optimization problem
with broad significance in operations research and computer science. Given a complete weighted graph, the
goal is to find a minimum-cost Hamiltonian cycle that visits each city exactly once and returns to the
starting city. Since no polynomial-time algorithm is known for solving arbitrary instances optimally, a
large spectrum of exact and heuristic methods has been developed. Importantly for this study, TSP solver
performance is typically \emph{heterogeneous}: which solver performs best can depend strongly on instance
structure (e.g., spatial patterns) and on the available runtime budget, making TSP a natural testbed for
algorithm selection.

\paragraph{Exact vs.\ heuristic algorithms.}
\emph{Exact} algorithms (e.g., branch-and-cut) guarantee optimality but can be computationally expensive as
instance size grows. The state-of-the-art exact solver \emph{Concorde} combines cutting-plane techniques
with branch-and-bound/branch-and-cut to solve instances to proven optimality and is widely used as a
reference baseline \citep{applegate2006tsp}. In contrast, \emph{heuristics} and metaheuristics trade
optimality guarantees for scalability and speed, aiming to obtain high-quality tours within practical time
limits. In many applications---and in budgeted evaluation settings such as ours---the relevant question is
not whether a solver can eventually prove optimality, but which method delivers the best solution
\emph{under a given cutoff}.

\begin{figure}[!t]
    \centering
    \includegraphics[width=0.5\linewidth, trim=214 116 188 78, clip]{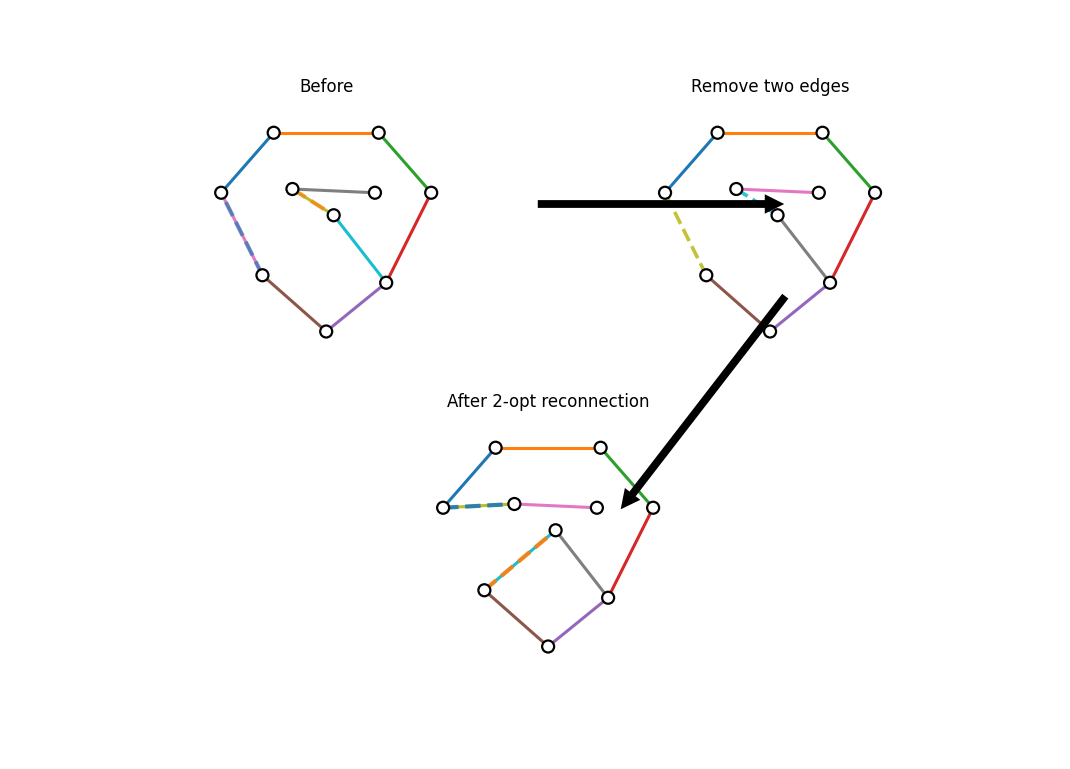}
    \caption{An example of a 2-opt move: remove two edges and reconnect the paths.}
    \label{fig:2opt_example}
\end{figure}

\paragraph{Heuristics and metaheuristics.}
Many high-performing TSP heuristics are built around iterative local improvement. A canonical operation is
\emph{2-opt} (Figure~\ref{fig:2opt_example}), which removes two edges and reconnects the resulting paths
to reduce tour length. Such operators form the foundation of the Lin--Kernighan (LK) heuristic, which
generalizes $k$-opt exchanges with variable depth. Its highly optimized variant, the
\emph{Lin--Kernighan--Helsgaun (LKH)} algorithm, is widely regarded as one of the strongest inexact TSP
solvers due to carefully engineered candidate sets and efficient search control \citep{helsgaun2000lkh}.
Beyond local search, population-based metaheuristics introduce diversification mechanisms that help escape
local optima. For example, Nagata and Kobayashi’s \emph{Edge Assembly Crossover (EAX)} is a powerful
genetic algorithm that recombines tours while preserving strong substructures \citep{nagata2012eax}, while
the \emph{Multiagent Optimization System (MAOS)} leverages cooperative multi-agent search to balance
exploration and exploitation \citep{xie2008maos}. A key empirical observation is that these methods are
often \emph{complementary}: some improve rapidly early in the run, while others refine more effectively
given additional time; moreover, their relative strengths vary across instance families. This
complementarity underpins the potential benefit of per-instance, budget-aware algorithm selection.

\paragraph{Learning-based approaches}
Recent work increasingly explores machine learning (ML) for TSP, either to construct tours directly or to
guide classical search. Early work using \emph{Pointer Networks} trained with reinforcement learning
demonstrated that neural models can produce near-optimal tours on certain Euclidean distributions, and
subsequent attention-based and transformer-style architectures further improved performance in
data-generated settings. Other approaches use learning to guide or accelerate local search (e.g., learning
adaptive $k$-opt policies). Among ML paradigms, \emph{Graph Neural Networks (GNNs)} are particularly
well suited to TSP because they operate directly on graph-structured data and can learn embeddings that
reflect geometric and structural properties of instances. While purely learning-based solvers do not
consistently outperform the strongest handcrafted heuristics across standard benchmarks, ML is
increasingly effective as a \emph{meta-level tool}---for example, to decide \emph{which} solver to apply
to a given instance under a fixed computational budget.

\paragraph{Algorithm selection and TSP benchmarks.}
TSP is a natural domain for algorithm selection (AS) due to the diversity of available solvers and their
instance-dependent performance \citep{kotthoff2015improving}. Public benchmark libraries such as
\emph{TSPLIB} provide standardized instances across a wide range of sizes and structural characteristics,
enabling systematic empirical comparison. Empirical studies further show that solver dominance may shift
with the runtime budget: under short cutoffs, ``fast improvers'' can be preferred, while under longer
budgets, solvers with stronger long-run refinement may dominate. This budget sensitivity motivates
selection policies that are trained and evaluated \emph{at the target cutoff}, and learning objectives that
emphasize \emph{relative solver ordering} (selection quality) rather than only absolute cost prediction.
In this work, we follow this budgeted setting by training and testing selectors under fixed cutoffs (10s
and 60s) and analyzing how the induced solver performance structure changes across budgets.

\paragraph{TSP Algorithm Portfolio}

We consider a portfolio of strong and diverse TSP solvers, following the anytime TSP-AS benchmark
setting in \citep{huerta2022aaas}. The portfolio spans distinct search biases (fast local improvement,
population-based diversification, cooperative search, and exact optimization), which is important because
selection gains rely on complementary strengths rather than near-duplicate solvers.

\begin{itemize}[leftmargin=*]
    \item \textbf{Chained Lin--Kernighan (CLK)} is an iterative local search method based on the LK
    heuristic, implemented in the Concorde library \citep{applegate2003clk}. It alternates LK-style search
    with perturbations (e.g., double-bridge moves) to escape local minima, making it a fast baseline that
    often achieves strong early improvements.

    \item \textbf{Lin--Kernighan--Helsgaun (LKH)} is a highly optimized LK variant with advanced candidate
    selection and move strategies, offering a strong trade-off between speed and solution quality
    \citep{helsgaun2000lkh}. It is widely considered a state-of-the-art heuristic for many Euclidean TSP
    instances.

    \item \textbf{Edge Assembly Crossover (EAX)} is a genetic algorithm specialized for TSP that uses an
    edge-assembly recombination operator to preserve high-quality substructures while enabling effective
    diversification \citep{nagata2012eax}. EAX is known to be particularly competitive on many benchmark
    families given sufficient runtime.

    \item \textbf{Multiagent Optimization System (MAOS)} is a population-based metaheuristic in which
    multiple agents perform local search and interact through a shared environment to balance exploration
    and exploitation \citep{xie2008maos}. Its cooperative dynamics can yield robust performance on
    heterogeneous instance sets.

    \item \textbf{Concorde} is an exact branch-and-cut solver that achieves provable optimality by solving
    a sequence of strengthened linear relaxations with cutting planes (e.g., subtour elimination and comb
    inequalities) integrated into a branch-and-bound framework \citep{applegate2006tsp}. Although expensive
    on large instances, it provides a gold-standard reference and also supplies strong heuristic
    components (including CLK) for primal solutions.
\end{itemize}

\paragraph{Connection to our model.}
Because the top best solvers can vary across instances and budgets, our GNN-based selector learns instance representations directly from city coordinates and edge distances and is trained with objectives that prioritize correct solver ordering. The above portfolio therefore provides a realistic and diverse selection space for studying budgeted, learning-based AS on TSP.

\subsection{Algorithm Selection: Problem Formulation}

Algorithm Selection (AS), first formalized by Rice~\citep{rice1976algorithm}, addresses the problem of selecting the most suitable algorithm for a given problem instance. Let $\mathcal{I}$ denote the instance space, $\mathcal{P}={A_1,\dots,A_m}$ a portfolio of candidate algorithms, and $\rho:\mathcal{P}\times\mathcal{I}\rightarrow\mathbb{R}$ a performance measure (e.g., runtime or solution cost). The goal of AS is to learn a selector $S:\mathcal{I}\rightarrow\mathcal{P}$ that minimizes the expected performance:
\[
\mathbb{E}_{i\sim\mathcal{D}}[\rho(S(i), i)],
\]
where $\mathcal{D}$ is a distribution over instances.

Two important reference points are the \emph{virtual best solver} (VBS), defined as
\[
S_{\mathrm{VBS}}(i)=\arg\min_{A\in\mathcal{P}}\rho(A,i),
\]
and the \emph{single best solver} (SBS), which is the best fixed algorithm across all instances. The objective of AS is to bridge the gap between SBS and VBS by exploiting instance-dependent performance differences. A typical AS workflow -instance representation, performance estimation and solver selection - is illustrated in Figure~\ref{fig:alg-selection}

\begin{figure}[!ht]
\centering
\includegraphics[width=1\textwidth]{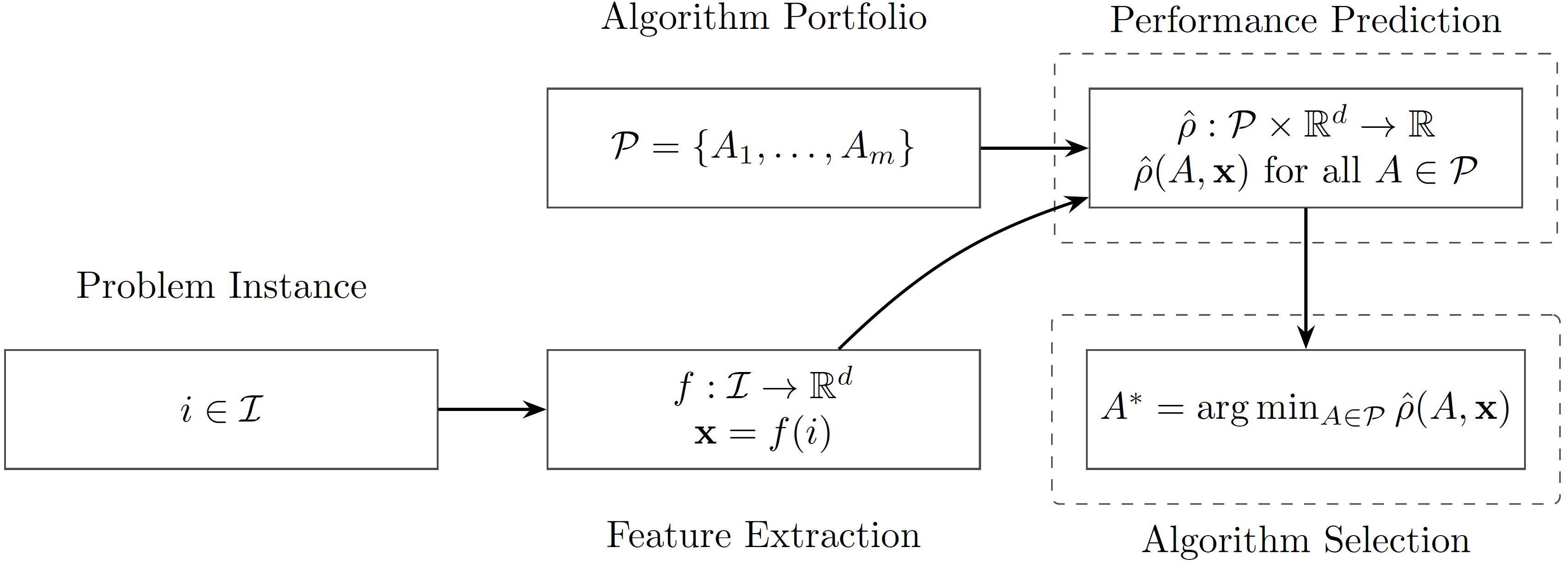}
\caption{A standard Algorithm Selection (AS) pipeline: represent the instance, predict relative solver
performance, and select the most promising algorithm.}
\label{fig:alg-selection}
\end{figure}

In practice, AS is implemented through performance prediction, classification, or learning-to-rank formulations~\citep{Kotthoff2014AlgorithmSelection,xu2008satzilla}. Since the ultimate goal is correct solver selection rather than precise cost estimation, ranking-based objectives are often more appropriate, especially when solver performances are close.

\subsection{Learning-based Algorithm Selection}

Traditional AS approaches rely on handcrafted instance features and supervised learning models~\citep{Kotthoff2014AlgorithmSelection,bischl2016aslib}. While effective, such methods depend heavily on feature quality and may not generalize well across domains.

To address this limitation, recent work explores representation learning for AS. Early approaches include image-based representations processed by convolutional neural networks~\citep{loreggia2016deep}, but these often fail to preserve the intrinsic graph structure of combinatorial problems. More recent methods employ graph neural networks (GNNs), which directly operate on graph-structured data and learn expressive instance embeddings~\citep{scarselli2008graph,wu2020comprehensive,prates2019learning}.

For TSP specifically, spatial-information-enhanced GNN models capture geometric relationships between cities without requiring handcrafted features~\citep{song2025ginestspas}. Extensions incorporating geometric invariance further improve robustness and generalization~\citep{song2026iegnntspas}. These developments demonstrate the effectiveness of GNNs for learning instance representations in algorithm selection.

Beyond algorithm selection, learning-based adaptation has also been studied within optimization algorithms. For example, meta-learning approaches can dynamically adjust search operators during optimization using sequence models such as LSTMs~\citep{jeong2025meta}. While these methods improve the internal dynamics of a single solver, they differ from AS, which operates at the level of selecting among multiple solvers.

\subsection{Traveling Salesman Problem and Solver Diversity}

The Traveling Salesman Problem (TSP) is a canonical NP-hard problem that seeks the shortest Hamiltonian cycle in a weighted graph. Due to its computational complexity, a wide range of exact and heuristic algorithms have been developed.

Exact solvers such as Concorde guarantee optimality but are computationally expensive for large instances~\citep{applegate2006tsp}. In contrast, heuristic and metaheuristic methods trade optimality for efficiency. Local search methods, such as Lin--Kernighan--Helsgaun (LKH), achieve strong performance through iterative improvement~\citep{helsgaun2000lkh}. Population-based approaches, such as Edge Assembly Crossover (EAX) and Multi-Agent Optimization System (MAOS), provide enhanced exploration capabilities~\citep{nagata2012eax,xie2008maos}. Chained Lin--Kernighan (CLK) combines local search with perturbation strategies to escape local optima~\citep{applegate2003clk}.

These solvers exhibit complementary strengths, and their relative performance can vary significantly across instances and runtime budgets. This heterogeneity makes TSP a natural testbed for algorithm selection.

Beyond the classical formulation, variants such as the Traveling Salesman Problem with Drone (TSP-D)~\citep{lee2025iterative} and asymmetric TSP~\citep{guettala2026heterogeneous} further highlight the diversity and complexity of routing problems, reinforcing the need for adaptive solution strategies.

\subsection{Learning-based and Hybrid Methods for TSP}

Recent advances in machine learning have led to a growing body of work on learning-based TSP solvers. Neural approaches include constructive models that generate tours sequentially and improvement methods that refine existing solutions. While promising, purely neural methods often struggle to generalize to large-scale or real-world instances.

Hybrid approaches that combine learning with classical heuristics have therefore gained increasing attention. Neural-guided local search methods \citep{hudson2021graph} enhance traditional optimization by learning decision policies within the search process. Similarly, neural-enhanced metaheuristics, such as neural-focused ant colony optimization (NeuFACO), incorporate learned guidance into classical search mechanisms~\citep{dat2025neufaco}.

Several recent works focus on scalability and generalization. DualOpt employs a divide-and-optimize strategy to handle large-scale TSP instances~\citep{zhou2025dualopt}. UNiCS integrates learning into both local search and population-based search, achieving strong performance across instance sizes~\citep{lv2025cascaded}. GELD proposes a unified neural architecture capable of solving TSP instances across different scales without retraining~\citep{xiao2025geld}, while ViTSP leverages pretrained vision-language models to guide optimization~\citep{yin2025vitsp}.

In addition to improving single-solution quality, alternative directions have been explored. Diversity optimization methods aim to generate multiple high-quality solutions~\citep{li2025diversity}, while hardware-inspired approaches such as probabilistic greedy solvers based on magnetic tunneling junctions introduce stochastic optimization mechanisms~\citep{zhang2025probabilistic}. These developments illustrate the diversity of modern TSP solution strategies.

\subsection{Algorithm Selection for TSP}

Building on these advances, algorithm selection for TSP has gained increasing attention. Early approaches rely on handcrafted features and classical machine learning models, while more recent methods adopt deep learning-based representations.

CNN-based approaches represent TSP instances as images, but this transformation may introduce irrelevant information and fail to capture structural properties~\citep{huerta2022aaas}. In contrast, GNN-based methods directly model graph structure and have shown improved performance~\citep{song2025ginestspas,song2026iegnntspas}.

However, existing TSP-AS studies often consider only small solver portfolios and formulate the problem as classification. Such formulations may not scale well and fail to capture richer relationships between solvers, such as relative performance rankings. Moreover, evaluation is typically limited to accuracy, without considering cost-based or rank-based metrics.

\subsection{Research Gap and Motivation}

Despite significant progress, several limitations remain in current TSP-AS research. First, most studies consider only a limited number of solvers, restricting their applicability in realistic scenarios. Second, classification-based formulations overlook relative performance differences among solvers. Third, the interaction between runtime budgets and solver performance is rarely analyzed in depth.

In this work, we address these limitations by proposing a GNN-based algorithm selection framework for TSP. Our approach learns instance representations directly from graph data and models solver performance using both cost and rank information. We evaluate the framework on a diverse solver portfolio under multiple runtime budgets and provide comprehensive analysis of algorithm selection behavior.

\end{hidden}

\section{Background}
\label{sec:background}


\subsection{Algorithm Selection (AS)}
\label{sec:algosel}

Algorithm Selection (AS) \citep{rice1976algorithm,kerschke2019automated} addresses the fact that algorithm performance is often instance-dependent: different solvers perform best on distinct instances so no single algorithm is uniformly superior across all instances. 
To formalize,
let $\mathcal{I}$ denote an instance space, $\mathcal{P}=\{A_1,\dots,A_m\}$ a portfolio of candidate algorithms, and $\rho:\mathcal{P}\times\mathcal{I}\rightarrow\mathbb{R}$ a performance measure such as runtime or solution cost. The goal is to learn a selector $S$ that minimizes expected performance over an instance distribution $\mathcal{D}$:
\[
\mathbb{E}_{i\sim\mathcal{D}}\left[\rho\!\left(S(i),\,i\right)\right].
\]

Two standard reference performances come from the \emph{Oracle} / \emph{Virtual Best Solver} (VBS) which chooses the best-performing algorithm for each instance, and the \emph{Single Best Solver} (SBS) which is the best single standalone solver concerning all the instances. 
In other words, VBS indicates the AS performance limit.
AS aims to primarily outperform SBS and bridge the gap between SBS and VBS by exploiting regularities in the relationship between instance characteristics and algorithm's performances. 
A typical and traditional AS workflow is illustrated in Figure~\ref{fig:alg-selection}.

\begin{figure}[!ht]
\centering
\includegraphics[width=1\textwidth]{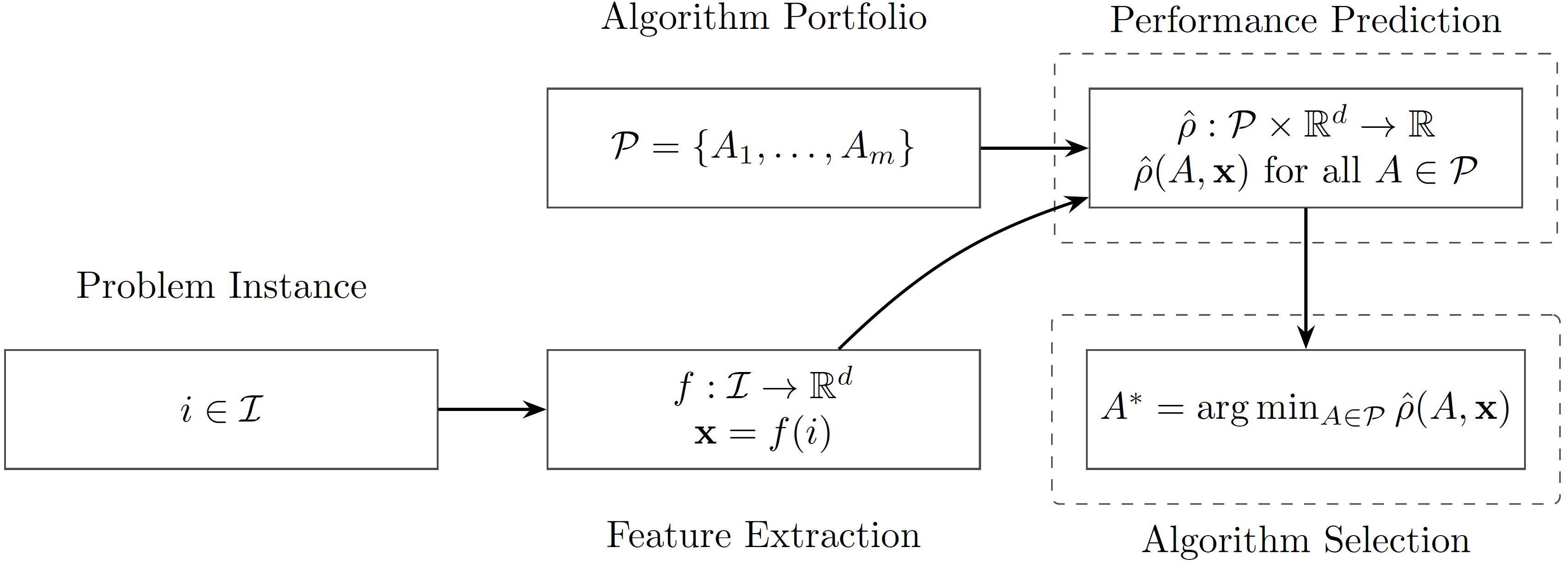}
\caption{A standard Algorithm Selection (AS) pipeline: represent the instance, predict relative solver performance, and select the most promising algorithm.}
\label{fig:alg-selection}
\end{figure}

Classical AS systems largely rely on hand-crafted instance features to feed supervised models. 
To exemplify, ASlib \citep{bischl2016aslib} established a standardized benchmark library containing different AS scenarios for various problems, where each scenario relies on predefined, hand-picked features that dictate the quality of the resulting models. 
Because these features are highly problem or domain-specific, building effective models tend to require costly, expert-level feature engineering. 
To reduce this dependence, one path is exploring representation learning. An initial step in this direction involves latent-factor models, which still use hand-crafted instance features but map them to a latent space reflecting candidate algorithm performance \citep{misir2017alors}.
To fully avoid explicitly designed instance features, Convolutional Neural Networks (CNNs) have emerged as a viable choice by utilizing simplified visual representations of the instances, as seen in Boolean Satisfiability (SAT) \citep{loreggia2016deep} and black-box function optimization \citep{yuan2026beyond}. More recently, leveraging the representational power of Transformers, embedding models have been developed to offer rich features that support strong AS models \citep{wang2026molecular}. 
However, for graph-structured optimization problems, GNN-based AS offers the most natural alternative for learning instance characterization directly from graphs. Because GNNs propagate information through nodes and edges via message passing \citep{scarselli2008graph,wu2020comprehensive}, they are especially suitable for these structures. 
Recent GNN-based AS studies in domains such as molecular docking further demonstrate that learned graph representations can successfully support budget-aware and multi-criteria solver selection without relying on manually engineered features \citep{ym2024gnnasdock,cao2025mcgnnasdock}.


\subsection{Traveling Salesman Problem (TSP)}
\label{subsec:tsp}


The Traveling Salesman Problem (TSP) is a fundamental NP-hard combinatorial optimization problem seeking the minimum-cost Hamiltonian cycle in a weighted graph. 
Beyond its theoretical significance, TSP serves as a foundational problem for numerous practical applications such as vehicle routing, circuit board drilling and VLSI chip design.
Thus, advancements in TSP algorithms are benefited in these diverse domains. 
While exact solvers such as Concorde may optimally solve certain instances with tens of thousands of cities, their exponential worst-case complexity make them computationally impractical for larger instances \citep{applegate2006tsp}.
This fact motivates to develop meta/-heuristic approaches for solving TSP under practical time budgets. 
Standard benchmark libraries like TSPLIB \citep{reinelt1991tsplib}, have played a central role in evaluating these solvers and establishing reproducible empirical comparisons.


Concerning the TSP algorithms, among the exact ones, Concorde remains the gold standard by combining branch-and-cut with highly effective cutting planes. 
For the heuristic variants, the most successful are local-search-based methods such as Lin--Kernighan--Helsgaun (LKH) which refines tours through sophisticated variable-depth edge exchanges \citep{helsgaun2000lkh}, and Chained Lin--Kernighan (CLK) which augments LK-style search with perturbation mechanisms to escape local optima \citep{applegate2003clk}. 
Furthermore, population-based approaches such as Edge Assembly Crossover (EAX) \citep{nagata2012eax} and the Multiagent Optimization System (MAOS) \citep{xie2008maos} emphasize diversification and cooperative exploration, respectively. 
Consequently, these solvers exhibit complementary strengths: some yield rapid improvements under short time budgets while others benefit more from extended search periods or specific instance structures.

Recently, the focus has shifted toward exploring learning-based and hybrid approaches for the TSP. 
The relevant reviews have highlighted the expanding role of Machine Learning (ML) in this domain~\citep{alanzi2025solving} while specific studies demonstrate that learned guidance can significantly improve local search~\citep{hudson2021graph}. 
Building on this direction, several large-scale methods have been developed to address scalability from different perspectives: DualOpt~\citep{zhou2025dualopt} employs divide-and-optimize decomposition, UNiCS~\citep{lv2025cascaded} combines learned guidance with cascaded local and population-based search, GELD~\citep{xiao2025geld} targets cross-scale neural generalization, and ViTSP~\citep{yin2025vitsp} integrates vision-language-model guidance into TSP optimization. Collectively, these studies indicate that ML is increasingly valuable not only for direct solution construction but also as a meta-level component complementing classical solvers.

\subsection{Algorithm Selection for TSP}
\label{sec:tsp_as_gap}
The TSP has long been recognized as an ideal domain for AS due to its rich solver ecosystem and strong instance-dependent performance variability \citep{kotthoff2015improving}. 
Earlier AS studies for TSP and related problems mainly relied on hand-crafted features and traditional machine learning models \citep{wagner2018case,misir2022algorithm,heins2023study}. 
While these studies demonstrated that per-instance selection can outperform a single static solver, they inherently suffered from the limitations of manual feature engineering.

Recent advancements have explored deep learning for TSP-specific AS. 
For instance, one framework formulated TSP-AS in an anytime setting by converting instances into image-like representations, allowing a Convolutional Neural Network (CNN) to predict solver behavior over time \citep{huerta2022aaas}. 
Although this successfully framed the problem for anytime AS, the method relies on intermediate visual encodings rather than exploiting the native graph structure of the TSP. 
To address this, a Graph Neural Networks (GNN)-based approach of GINES \citep{song2025revisit} was introduced, enabling TSP instances to be processed directly as graphs. 
This avoids manual feature design and image generation; however, the approach remains restricted to binary selection between only two candidate algorithms. 
Further extending this line of work, geometric invariance and equivariance were incorporated into GNN-based TSP-AS and hardness prediction, demonstrating that geometry-aware models improve robustness under Euclidean transformations \citep{song2025geometrically}.

Despite this progress, several critical limitations remain in current TSP-AS research. Existing GNN-based studies predominantly consider binary selection settings, formulate AS primarily as a classification problem, and evaluate performance almost exclusively through accuracy \citep{song2025revisit,song2025geometrically}. 
Such formulations are less suitable for larger and realistic portfolios, especially when multiple solvers deliver relatively close performances. 
Moreover, the interaction between runtime budgets and solver performance structures has received limited attention, even though budget sensitivity is central to practical TSP solving.

In the present work, we address these limitations by proposing a GNN-based AS framework for TSP that operates on a larger, more diverse solver portfolio, predicts per-solver costs under fixed computational budgets, and investigates both cost-based and rank-based learning objectives. 
This formulation enables evaluation beyond simple classification accuracy, providing a more informative view of selection quality under varying runtime constraints.

\section{Methodology}
\label{sec:methodology}


\subsection{Dataset and Data Preprocessing}

\begin{figure}[htbp]
\centering
\begin{tikzpicture}[
    scale=0.91, transform shape,
    x=1cm, y=1cm,
    block/.style={rectangle, draw=dark!30, fill=white, rounded corners=4pt, 
                  minimum width=2.8cm, minimum height=1.4cm, align=center, 
                  font=\small\sffamily\color{dark}, inner sep=6pt},
    list_block/.style={rectangle, draw=dark!30, fill=white, rounded corners=4pt, 
                       minimum width=2.8cm, minimum height=2.4cm, align=center, 
                       font=\footnotesize\sffamily\color{dark}, inner sep=6pt},
    math/.style={font=\footnotesize\itshape, text=dark!70, align=center},
    arrow/.style={-{Latex[length=2.5mm, width=2mm]}, thick, dark!60},
    container/.style={rectangle, draw=dark!15, fill=lightbg, rounded corners=6pt},
    containerSecond/.style={rectangle, draw=dark!15, fill=lightSoftGreen, rounded corners=6pt}
]
\definecolor{accent}{RGB}{0, 114, 189}
\definecolor{dark}{RGB}{33, 37, 41}
\definecolor{lightbg}{RGB}{248, 249, 250}
\definecolor{lightSoftGreen}{RGB}{247, 250, 242}   

    \node[container, minimum width=17.5cm, minimum height=4.8cm] at (8, 5.5) {};
    \node[font=\small\sffamily\bfseries\color{dark}, rotate=90, anchor=east] at (8.3 - 17.5/2, 6.2) {Dataset};
    
    \node[containerSecond, minimum width=17.5cm, minimum height=5.2cm] at (8, 0) {};
    \node[font=\small\sffamily\bfseries\color{dark}, rotate=90, anchor=east] at (8.3 - 17.5/2, 1) {Metasolver};

    
    \node[list_block] at (2, 5.5) {
        \textbf{Generated Instances} \\[0.1cm]
        (\textbf{Train}) \\[0.3cm]
        RUE \\ NETGEN \\ NETGENM \\ TSPGEN
    };
    
    \node[list_block] at (5.9, 5.5) {
        \textbf{Public Instances} \\[0.1cm]
        (\textbf{Test}) \\[0.3cm]
        TSPLIB \\ NATIONAL \\ VLSI \\ TNM
    };    
    
    \node[list_block] at (9.5, 5.5) {
        \textbf{Solvers} \\[0.3cm]
        CONCORDE \\ CLK \\ EAX \\ LKH \\ MAOS
    };
    
    \node[block, fill=accent!5, minimum height=2.4cm] at (14.5, 5.5) {
        \textbf{Cost and Time} \\[0.2cm]
        \footnotesize Improving solution \\ trajectories
    };

    \node at (7.75, 5.5) [font=\large\color{dark!60}] {$\times$}; 
    \draw[->, thick, dark!60] (10.9, 5.5) -- (12.9, 5.5);

    
    \node[block] at (1.75, 1.0) {
        \textbf{Input} \\[0.2cm]
        \footnotesize Node Coordinates \\ $\mathbf{X} \in \mathbb{R}^{N \times 2}$
    };
    
    \node[block, fill=accent!5, minimum width=3.5cm, minimum height=1.8cm] at (6.3, 1.0) {
        \textbf{ResGatedGCN Model} \\[0.2cm]
        \footnotesize $\mathbf{H}^{(L)} = f_{\theta}(\mathbf{X}, \mathbf{A}, \mathbf{E})$
    };
    
    \node[block, minimum width=3.5cm, minimum height=1.8cm] at (12.7, 1.0) {
        \textbf{Labels} \\[0.2cm]
        \footnotesize Cost (standardized) and rank of each solver \\ $\mathbf{y} \in \mathbb{R}^m$
    };
    
    \node[block, fill=accent!15] at (6.3, -1.6) {
        \textbf{Output} \\[0.2cm]
        \footnotesize Predicted Costs: $\hat{\mathbf{y}} \in \mathbb{R}^m$
    };

    \draw[->, thick, dark!60] (3.3, 1.0) -- (4.215, 1.0); 
    \draw[->, thick, dark!60] (9.26, 1.0) -- (8.39, 1.0); 
    \draw[->, thick, dark!60] (6.35, 0.1) -- (6.35, -0.9); 

    
    \draw[->, thick, dark!60] (8, 3.1) -- (8, 2.6); 
    
    \draw[->, thick, dark!60]  (14.5, 4.3) -- (14.5, 2.85) -| (12.5, 2);

\end{tikzpicture}
\caption{The high-level setup of the GNNAS-TSP framework. The \textbf{Dataset} module generates performance/cost and execution time values by running the listed TSP solvers on instances from the listed TSP benchmark sets. The \textbf{Metasolver} module uses a ResGatedGCN to map input node coordinates $\mathbf{X}$ to predicted solver costs $\hat{\mathbf{y}} \in \mathbb{R}^m$, trained against standardized cost and/or rank labels $\mathbf{y} \in \mathbb{R}^m$ for the solver portfolio $\mathcal{P} = \{A_1, \dots, A_m\}$.}
\label{fig:general_schema}
\end{figure}
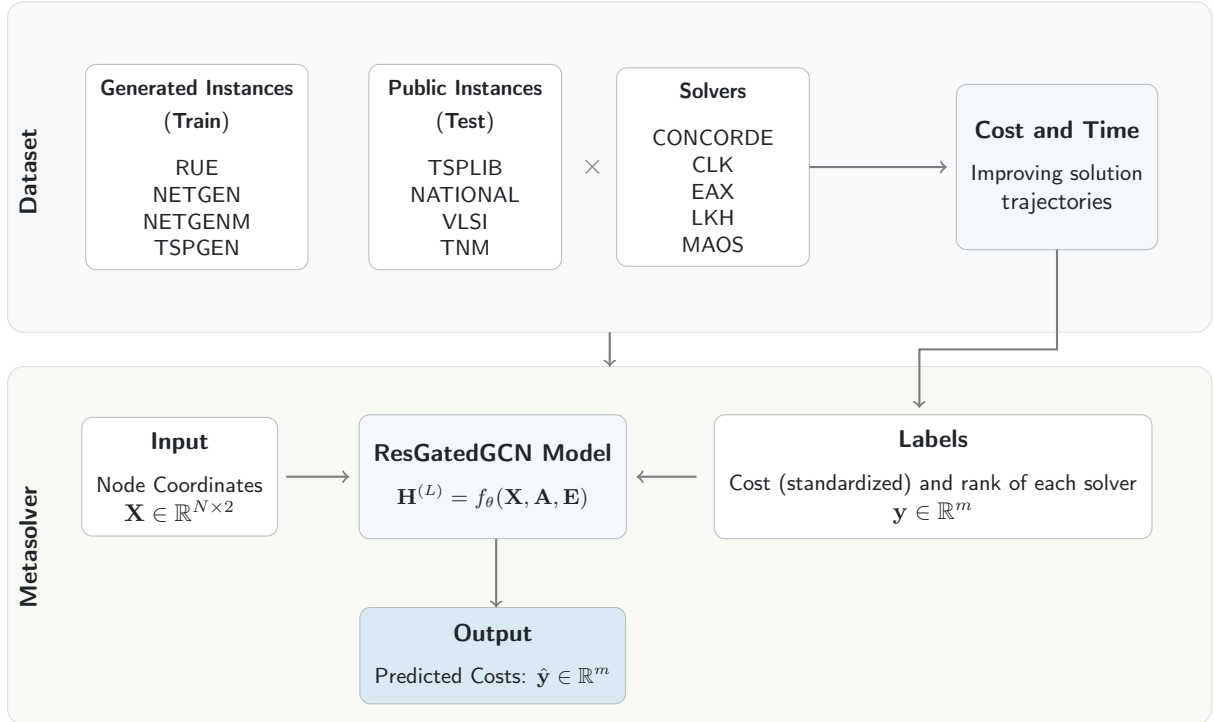

Figure ~\ref{fig:general_schema} summarizes the general schema of this work. 
The dataset used, introduced by \citet{huerta2022aaas}, implements an anytime Algorithm Selection (AS) framework. 
We directly use the training and test splits provided in that benchmark and do not generate any new problem instances. 
The training set consists of instances drawn from several instance generators, including Random Uniform Euclidean (RUE), NETGEN, NETGENM, and TSPGEN, which produce instances with diverse spatial structures and varying levels of difficulty. 
The testing set comprises public benchmark instances, including TSPLIB, TNM, NATIONAL, and VLSI. 
The algorithm portfolio comprises 5 well-known TSP solvers including Concorde, CLK, EAX, LKH and MAOS, all are briefly described in Section~\ref{subsec:tsp}.
Referring to \citet{huerta2022aaas}, the data here is extracted from the performance traces of each solver on the aforementioned TSP instances. 
Specifically, the raw data consists of the trajectory of the incumbent best solution values over time, recorded in seconds.

In our experiments, we adopt a fixed cutoff approach, with time budgets of 10 seconds and 60 seconds. 
For each instance, we choose the timestep which is smaller than and closest to the given time budget, and use its corresponding solution cost as the regression target. 
Since the TSP instances have different sizes and raw tour costs vary, we use normalized solver costs as learning targets. For each instance, the costs of the 5 candidate solvers are standardized relative to one another. 
Let $c_{ia}$ denote the solution cost obtained by solver $a \in \{1,\ldots,5\}$ on instance $i$ under a given time budget. We transform each cost as
\[
\tilde{c}_{ia}
=
\frac{c_{ia}-\mu_i}{\sigma_i+\varepsilon},
\]
where
\[
\mu_i = \frac{1}{5}\sum_{a=1}^{5} c_{ia},
\qquad
\sigma_i =
\sqrt{
\frac{1}{5}\sum_{a=1}^{5}
\left(c_{ia}-\mu_i\right)^2
},
\]
and $\varepsilon = 10^{-8}$ ensures numerical stability. 
Thus, if all candidate solvers obtain the same cost for an instance, all standardized targets are set to zero. This allows the model to focus on predicting the relative performance of solvers on the same instance rather than the absolute scale of the tour length. At prediction time, we select the solver with the lowest predicted standardized cost. The standardization is performed independently for each instance and for each time-budget setting.

Since rank-based losses are also considered during training, we construct rank labels for the candidate solvers. Solvers are ordered primarily by their observed solution cost under the fixed time budget, with runtime used only as a tie-breaker to produce a strict ordering for rank-based loss computation. These training rank labels are used for losses such as RankNet and LambdaRank.

This training rank should be distinguished from the rank metric used in evaluation. In the experimental analysis, we report average tied rank based on the observed final solution costs under the fixed time budget. Solvers with the same final cost are assigned the average of their occupied ranks. This evaluation metric avoids artificially separating solvers that achieve the same solution quality.

\subsection{GNNAS-TSP}

\begin{figure*}[htbp]
\centering
\begin{tikzpicture}[
    x=1.3cm, y=1.3cm,
    neuron/.style={circle, draw=dark, fill=white, minimum size=10pt, inner sep=0pt},
    graph_node/.style={circle, draw=dark, fill=accent!20, minimum size=12pt, inner sep=0pt},
    vector/.style={rectangle, draw=dark, fill=accent!10, minimum width=6pt, minimum height=20pt, inner sep=0pt},
    block/.style={rectangle, draw=dark!30, fill=white, rounded corners=3pt, 
                  minimum width=2.2cm, minimum height=1.2cm, align=center, 
                  font=\small\sffamily\color{dark}, inner sep=4pt},
    arrow/.style={-{Latex[length=2.5mm, width=2mm]}, thick, dark!60},
    math/.style={font=\footnotesize\itshape, text=dark!70, align=center}
]
\definecolor{accent}{RGB}{0, 114, 189}
\definecolor{dark}{RGB}{33, 37, 41}

    
    \begin{scope}[shift={(0,1.6)}]
        \node[graph_node] (n1) at (0, 0.6) {};
        \node[graph_node] (n2) at (0.5, 0.2) {};
        \node[graph_node] (n3) at (0.3, -0.4) {};
        \node[graph_node] (n4) at (-0.3, -0.4) {};
        \node[graph_node] (n5) at (-0.5, 0.2) {};
        \draw[thin, dark!50] (n1)--(n2)--(n3)--(n4)--(n5)--(n1);
        \draw[thin, dark!30] (n1)--(n3) (n2)--(n4) (n3)--(n5);
        \node at (0, -1.1) [font=\small\sffamily\bfseries\color{dark}] {Input Graph};
        \node at (0, -1.6) [math] {$\mathcal{G} = (\mathbf{X}, \mathbf{A}, \mathbf{E})$};
    \end{scope}
    \draw[->, thick, dark!60] (0.8, 1.6) -- (1.9, 1.6);

    \node[block] at (3.0, 1.6) {
        \textbf{ResGatedGCN} \\
        \footnotesize Message Passing
    };
    \node at (3.0, 0.0) [math] {$\mathbf{H}^{(L)} = f_{\theta}(\mathbf{X}, \mathbf{A}, \mathbf{E})$};
    \draw[->, thick, dark!60] (4.1, 1.6) -- (5.325, 1.6);

    \begin{scope}[shift={(6.0,1.6)}]
        \node[vector] at (-0.6, 0) {};
        \node[vector] at (-0.3, 0) {};
        \node[vector] at (0, 0) {};
        \node[vector] at (0.3, 0) {};
        \node[vector] at (0.6, 0) {};
        \node at (0, -1.1) [font=\small\sffamily\bfseries\color{dark}] {Node Embeddings};
        \node at (0, -1.6) [math] {$\mathbf{H}^{(L)} \in \mathbb{R}^{N \times h}$};
    \end{scope}
    \draw[->, thick, dark!60] (6.7, 1.6) -- (7.9, 1.6);

    \node[block] at (9.0, 1.6) {
        \textbf{Global Pooling} \\
        \footnotesize Attention
    };
    \node at (9.0, 0.0) [math] {\hspace{-0.22cm}$\mathbf{h}_G = \text{READOUT}(\mathbf{H}^{(L)})$};
    
    \draw[->, thick, dark!60] (9.0, 1.1) -- (9.0, -0.77);

    
    \begin{scope}[shift={(9.0,-1.2)}]
        \node[vector, minimum height=32pt, fill=accent!20] at (0, 0) {};
        \node at (0, -1.1) [font=\small\sffamily\bfseries\color{dark}] {Graph Vector};
        \node at (0, -1.6) [math] {$\mathbf{h}_G \in \mathbb{R}^h$};
    \end{scope}
    \draw[->, thick, dark!60] (8.9, -1.2) -- (6.9, -1.2);

    \begin{scope}[shift={(5.3,-1.2)}] 
        \node[neuron] (m1_1) at (0, 0.5) {};
        \node[neuron] (m1_2) at (0, 0) {};
        \node[neuron] (m1_3) at (0, -0.5) {};
        \node[neuron] (m2_1) at (0.7, 0.35) {};
        \node[neuron] (m2_2) at (0.7, -0.15) {};
        \node[neuron] (m2_3) at (0.7, -0.65) {};
        \node[neuron] (m3_1) at (1.4, 0.15) {};
        \node[neuron] (m3_2) at (1.4, -0.35) {};
        
        \foreach \i in {1,2,3} {
            \foreach \j in {1,2,3} {
                \draw[thin, dark!20] (m1_\i) -- (m2_\j);
            }
        }
        \foreach \i in {1,2,3} {
            \foreach \j in {1,2} {
                \draw[thin, dark!20] (m2_\i) -- (m3_\j);
            }
        }
        \node at (0.7, -1.1) [font=\small\sffamily\bfseries\color{dark}] {MLP Decoder};
        \node at (0.7, -1.6) [math] {$\hat{\mathbf{y}} = \text{MLP}(\mathbf{h}_G)$};
    \end{scope}
    \draw[->, thick, dark!60] (5.1, -1.2) -- (3.3, -1.2);
    
    \begin{scope}[shift={(3.0,-1.2)}]
        \node[vector, fill=accent!30] at (-0.2, 0) {};
        \node[vector, fill=accent!30] at (0.2, 0) {};
        \node at (0, -1.1) [font=\small\sffamily\bfseries\color{dark}] {Predicted Costs/Ranks};
        \node at (0, -1.6) [math] {$\hat{\mathbf{y}} \in \mathbb{R}^m$};
    \end{scope}

\end{tikzpicture}
\caption{Model architecture. The input TSP graph $\mathcal{G} = (\mathbf{X}, \mathbf{A}, \mathbf{E})$ is encoded via ResGatedGCN to produce node embeddings $\mathbf{H}^{(L)}$. These are aggregated into a graph-level representation $\mathbf{h}_G$ via global attention pooling, which is then decoded by an MLP to predict the normalized costs or ranks $\hat{\mathbf{y}} \in \mathbb{R}^m$ for the TSP solver portfolio $\mathcal{P} = \{A_1, \dots, A_m\}$.}
\label{fig:model_architecture}
\end{figure*}
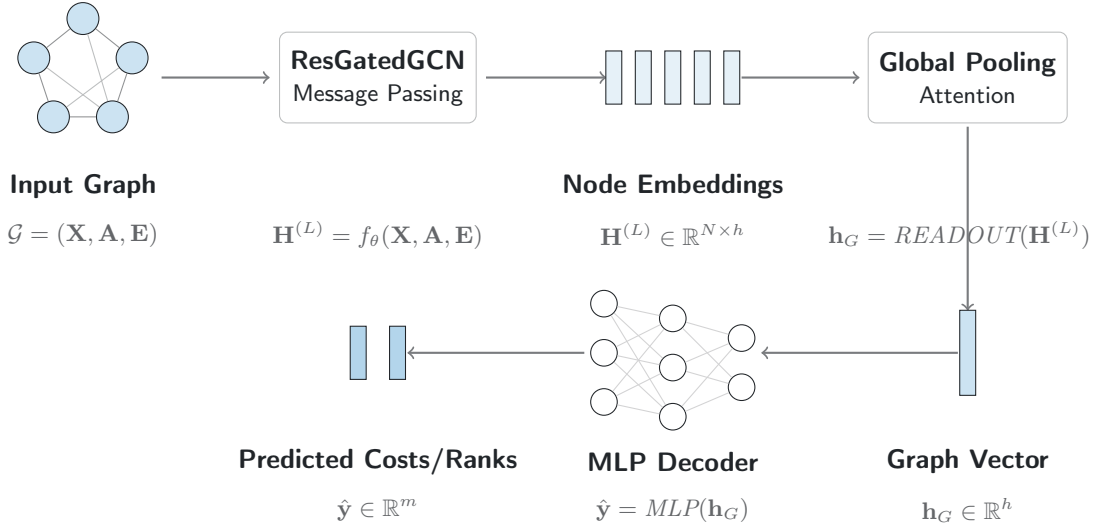

The overall workflow of our GNNAS-TSP's architecture is given in Figure ~\ref{fig:model_architecture}.  
The current architecture is built upon the Residual Gated Graph Convolutional Network (ResGatedGCN) from \citet{joshi2019rggcn}. 
The model takes as input a graph representation of a TSP instance $i \in \mathcal{I}$, denoted as $\mathcal{G} = (\mathbf{X}, \mathbf{A}, \mathbf{E})$, where $\mathbf{X} \in \mathbb{R}^{N \times 2}$ contains the 2-dimensional coordinates of $N$ cities, $\mathbf{A}$ is the adjacency matrix, and $\mathbf{E}$ represents edge features.
Each node coordinate is embedded into an $h$-dimensional initial feature vector using TransformerConv \citep{shi2020masked}, which is a GNN layer utilizing a transformer-style attention mechanism. 
The edge adjacency and distance matrices are respectively embedded into $\frac{h}{2}$-dimensional vectors using a fully-connected linear layer and concatenated to form the initial edge embeddings $\mathbf{E} \in \mathbb{R}^{|\mathcal{E}| \times h}$, where $\mathcal{E}$ denotes the set of edges and $|\mathcal{E}|$ is the total number of edges.

The adjacency matrix encodes edge types as follows: 0 for no connection, 1 for a regular edge, and 2 for a self-loop. In the current setting, we restrict our focus to complete graphs, which means each value of the adjacency matrix is either 1 or 2. This choice reflects the nature of TSP, where edges are possible between any pair of cities in a route. However, since optimal routes usually connect nodes with geographically close ones, k-nearest-neighbor (kNN) graphs may provide a more efficient alternative and are left for future investigation. Distances are calculated directly from node coordinates. A dense adjacency matrix is used for consistency with prior work. To control computational cost, we exclude instances with more than 1000 nodes. Sparse representation may be considered for scalability in future work.


Due to varying instance sizes, we process one graph per batch (batch size = 1). Batch normalization layers are kept as in the original implementation of \citet{joshi2019rggcn}, after convolution and before the activation function. Since each batch contains a single instance, the running statistics have limited effect.
The model updates node and edge embeddings using Residual Gated Graph Convolution layers, as proposed by \citet{joshi2019rggcn}. The node and edge features at layer $l+1$ are defined as:
\[
\mathbf{x}_i^{l+1} = \mathbf{x}_i^l + \text{ReLU}(\text{BN}(\mathbf{W}_1^l\mathbf{x}_i^l + \sum_{j \in \mathcal{N}(i)} \eta_{ij}^l \odot \mathbf{W}_2^l\mathbf{x}_j^l)) \text{ with } \eta_{ij}^l = \frac{\sigma(\mathbf{e}_{ij}^l)}{\sum_{j'\in\mathcal{N}(i)}\sigma(\mathbf{e}_{ij'}^l) + \epsilon}
\]
\[
\mathbf{e}_{ij}^{l+1} = \mathbf{e}_{ij}^l + \text{ReLU}(\text{BN}(\mathbf{W}_3^l\mathbf{e}_{ij}^l + \mathbf{W}_4^l\mathbf{x}_i^l + \mathbf{W}_5^l\mathbf{x}_j^l))
\]
where $\eta_{ij}^l$ is the edge gate that controls the contribution of neighboring nodes during message passing, and $\epsilon$ is a small numerical stability constant.

After $l_{\text{conv}}$ layers of message passing, the final node embeddings $\mathbf{H}^{(L)} \in \mathbb{R}^{N \times h}$ are extracted and aggregated using a Global Attention pooling operator to produce a single graph-level feature $\mathbf{h}_G \in \mathbb{R}^h$. This graph-level representation is then passed through $l_{\text{mlp}}$ layers of a Multi-Layer Perceptron (MLP) that outputs an $m$-dimensional vector $\hat{\mathbf{y}} \in \mathbb{R}^m$ for each instance, where each component $\hat{y}_k$ represents the predicted performance $\rho(A_k, i)$ for algorithm $A_k \in \mathcal{P}$. 
The learned selector $S$ is then defined as:
\[
S(i) = \arg\min_{k \in \{1,\dots,m\}} \hat{y}_k
\]
which selects the algorithm with the lowest predicted cost for instance $i$.

\subsection{Loss Function}
\label{subsec:loss function}

The models are trained via regression to predict algorithm costs under a fixed time budget. 
The primary objective of Algorithm Selection (AS) is to accurately identify the best or near-best solvers rather than to achieve absolute precision in the predicted cost values. 
We investigate two categories of loss functions to optimize the accuracy of cost predictions and the ordering of algorithms. 
We evaluate cost-based losses, rank-based losses and their combinations(0.5 cost loss + 0.5 rank loss). 
The cost-based ones maintain proximity between predicted and true solution costs. 
The rank-based losses encourage the model to correctly order the algorithms for each instance, especially the top ones.
Combining both losses guides the model to produce accurate magnitude predictions that effectively select the (near-)optimal solvers.

We employ Mean Squared Error (MSE), Mean Absolute Error (MAE), and Huber loss for cost-based optimization. 
MSE provides smooth gradients and penalizes large prediction errors strongly. 
MAE offers a constant gradient and measures the average absolute error. 
The Huber loss behaves quadratically for small errors and linearly for large errors, providing robustness against anomalies.
For rank-based loss functions, we accommodate pairwise, listwise, and metric-driven paradigms using RankNet, ListNet, and LambdaRank.

\textit{RankNet} formulates ranking as a pairwise classification problem \citep{burges2005ranknet}. 
For each pair of algorithms $k$ and $l$ with different training ranks, the predicted preference probability is defined as
\[
P_{kl} = \sigma(s_k-s_l),
\]
where $s_k = -\hat{y}_k$ and $s_l = -\hat{y}_l$ denote the predicted ranking scores for algorithms $k$ and $l$. We define the scores as the negative predicted costs because lower costs indicate better performance. Training ranks are determined primarily by observed solution cost under the fixed time budget; when two solvers obtain the same cost, the solver with the lower runtime is assigned the better rank. Consequently, RankNet also learns this secondary runtime preference for equal-cost pairs.
We define the scores as the negative predicted costs because lower costs indicate better performance. 
The function $\sigma(\cdot)$ is the sigmoid function. 
The ground-truth preference label is given by
\[
S_{kl} =
\begin{cases}
1 & \text{if } y_k < y_l, \\
0 & \text{otherwise},
\end{cases}
\]
where $y_k$ and $y_l$ are the true costs. 
The RankNet loss is defined as
\[
\mathcal{L}_{\mathrm{RankNet}} = -\frac{1}{|\mathcal{V}|} \sum_{(k,l)\in \mathcal{V}}
\Big[ S_{kl} \log P_{kl} + (1 - S_{kl}) \log (1 - P_{kl}) \Big],
\]
where $\mathcal{V}$ denotes the set of valid algorithm pairs.

\textit{ListNet} adopts a listwise formulation by modeling the predicted and ground-truth rankings as probability distributions over the algorithm list \citep{cao2007listnet}. For a batch element $b$, the normalized distributions are defined as
\[
p_k^{(b)} = \frac{\exp(-y_k^{(b)})}{\sum_{l=1}^{m} \exp(-y_l^{(b)})}, \quad
q_k^{(b)} = \frac{\exp(-\hat{y}_k^{(b)})}{\sum_{l=1}^{m} \exp(-\hat{y}_l^{(b)})},
\]
where $y_k^{(b)}$ and $\hat{y}_k^{(b)}$ denote the true and predicted costs of algorithm $k$ in batch element $b$, and $m$ is the number of algorithms in the portfolio. We construct the ListNet target distribution directly from ground-truth costs rather than discrete ranks to preserve relative cost differences and avoid information loss from rank discretization. The ListNet loss is the cross-entropy between these two distributions:
\[
\mathcal{L}_{\mathrm{ListNet}} = - \frac{1}{B} \sum_{b=1}^{B} \sum_{k=1}^{m} p_k^{(b)} \log q_k^{(b)},
\]
where $B$ is the batch size.

\textit{LambdaRank} is a metric-driven ranking approach that optimizes evaluation metrics such as NDCG by weighting pairwise losses according to their impact on the target metric \citep{burges2006lambdarank}. We adopt a LambdaRank-style surrogate objective defined as:
\[
\mathcal{L}_{\mathrm{Lambda}} = \frac{1}{|\mathcal{V}|} \sum_{(k,l)\in \mathcal{V}} \Delta \mathrm{NDCG}_{kl} \cdot \ell_{kl},
\]
where $\ell_{kl}$ denotes the RankNet pairwise logistic loss for algorithms $k$ and $l$, $\Delta \mathrm{NDCG}_{kl}$ represents the change in $\text{NDCG}@3$ induced by swapping their positions, and $\mathcal{V}$ is the set of valid algorithm pairs. The positions used in the $\Delta \mathrm{NDCG}$ computation are derived from the ground-truth rankings to ensure training stability and avoid fluctuations in pairwise weights caused by changing predicted orderings. 

LambdaRank is originally defined through gradient construction rather than an explicit loss function. We adopt a normalized surrogate objective to approximate its optimization behavior, normalizing the loss by the number of valid pairs to yield scale-invariant and stable training across different batch sizes and candidate set sizes.

\section{Experiments and Results}
\label{sec:results}
\subsection{Computational Environment}
\label{subsec:envir}

All experiments are conducted on a Linux-based server running Ubuntu~22.04.5~LTS. The server is equipped with 8 NVIDIA GeForce RTX~4090-series GPUs, each with 24~GB of memory. 
The NVIDIA driver version is~550.163.01, and the CUDA version is~12.4. 
The experiments are implemented in Python~3.10.18. 
All model variants and time-budget settings are evaluated under the same computational environment.
The complete setup, code and data are publicly available at \url{https://github.com/Bryan-Yang1/GNNAS-TSP}.

\subsection{Experimental Setup and Hyper-parameters}
\label{subsec:setup}

We use a filtered dataset containing 1000 instances in total, with 700 instances for training and 300 instances as the test set. 
The filtering here removes instances with missing or invalid solver outputs, incomplete labels for those 5 candidate solvers or more than 1000 nodes.
For each loss configuration, an Algorithm Selection (AS) model is built through that full set of 700 training instances. 

Concerning the explicit architectural and hyper-parameter details of GNNAS-TSP, 3 residual gated graph convolutional layers followed by a 2-layer MLP are used. 
The node feature dimension is set to $h=128$. 
Models are trained using the Adam optimizer with an initial learning rate of 0.001. 
We fixed this model architecture to a moderate GNN configuration following commonly used choices in related studies~\citep{kool2018attention,song2025revisit,joshi2019rggcn}. 
Table~\ref{tab:hyperparameters} summarizes these hyper-parameter settings. With the architecture and base optimizer settings fixed, we then performed a small one-factor sensitivity check for the learning-rate decay and the number of epochs. 
It revealed that a decay rate of 1.2 every 5 epochs and a maximum of 30 epochs make a suitable combination.

\begin{table}[ht]
\centering
\caption{Hyperparameter configuration of the GNNAS-TSP's architecture}
\begin{tabular}{l|c|l}
\toprule
\textbf{Hyperparameter} & \textbf{Value} & \textbf{Description} \\ \hline
Hidden dimension ($h$) & 128 & Feature dimension in all layers \\
Convolutional layers ($l_{conv}$) & 3 & Number of residual gated graph conv layers \\
MLP layers ($l_{mlp}$) & 2 & Layers in output prediction head \\
Learning rate & 0.001 & Initial learning rate for Adam optimizer \\
Decay rate & 1.2 & Learning rate divisor every 5 epochs \\
Max epochs & 30 & Total training epochs \\ 
\bottomrule  
\end{tabular}
\label{tab:hyperparameters}
\end{table}

The following analyses are based on predictions for the held-out 300-instance test set. We use the aggregate test-set results in Table~\ref{tab:overall_perf} to identify one representative high-performing loss configuration for each time budget. The subsequent VBS-gap, solver-selection behaviour, and SVD analyses characterize these selected configurations on the same test set. Accordingly, these analyses are post-selection and exploratory. The reported Wilcoxon $p$-values summarize paired differences within this test sample and are not interpreted as independent confirmatory inference following model selection.

\subsection{Overall Performance Evaluation}
\label{subsec:perf}

\begin{table}[htbp]
\centering
\caption{Aggregate performance summary for 10s and 60s runtime budgets. Lower cost and lower rank indicate better performance. The rank metric is the average tied rank computed from observed final solver costs under the fixed time budget; tied solvers receive the average of their occupied ranks.}
\label{tab:overall_perf}
\resizebox{\textwidth}{!}{
\begin{tabular}{llrrr | rrr}
\toprule
\multirow{2}{*}{\textbf{Budget}} & \multirow{2}{*}{\textbf{Method}} & \multicolumn{3}{c}{\textbf{Cost}} & \multicolumn{3}{c}{\textbf{Rank}} \\
\cmidrule(lr){3-5} \cmidrule(lr){6-8}
& & \textbf{Mean} & \textbf{Median} & \textbf{Std} & \textbf{Mean} & \textbf{Median} & \textbf{Std} \\
\midrule

\multicolumn{8}{l}{\textbf{10s budget}} \\
\midrule
10s & \textbf{VBS} & -0.5069 & -0.5172 & 0.2925 & 1.7617 & 1.5000 & 0.5746 \\
10s & \textbf{SBS}$_{\text{cost}}$ (\textbf{CLK}) & -0.4481 & -0.5036 & 0.2501 & 2.7333 & 3.0000 & 0.6785 \\
10s & \textbf{SBS}$_{\text{rank}}$ (\textbf{MAOS}) & -0.4440 & -0.5153 & 0.4724 &  1.9450 & 2.0000 & 0.7459 \\
\midrule
10s & LKH & -0.3664 & -0.5118 & 0.4904 & 2.1417 & 2.0000 & 0.8405 \\
10s & Concorde & -0.3100 & -0.4555 & 0.4687 & 3.4967 & 4.0000 & 0.8076 \\
10s & EAX & 1.6336 & 1.9988 & 0.8031 & 4.6833 & 5.0000 & 0.7659 \\

\midrule

10s & \cellcolor{gray!20} \textbf{MAE+LambdaRank} & \cellcolor{gray!20} \textbf{-0.4912} & -0.5148 & 0.3211 & 1.9483 & 2.0000 & 0.6554 \\
10s & MSE & -0.4803 & -0.5151 & 0.3556 & 1.9533 & 2.0000 & 0.6691 \\
10s & MAE+ListNet & -0.4786 & -0.5145 & 0.3782 & 1.9683 & 2.0000 & 0.6792 \\
10s & ListNet & -0.4761 & \cellcolor{gray!20} \textbf{-0.5158} & 0.3792 & \cellcolor{gray!20} \textbf{1.9350} & 2.0000 & 0.6818 \\
10s & MAE+RankNet & -0.4697 & -0.5136 & 0.3354 & 2.0167 & 2.0000 & 0.6680 \\
10s & MSE+ListNet & -0.4683 & -0.5145 & 0.3721 & 2.0083 & 2.0000 & 0.6921 \\
10s & Huber+RankNet & -0.4678 & -0.5145 & 0.3715 & 1.9933 & 2.0000 & 0.6891 \\
10s & MAE & -0.4660 & -0.5136 & 0.3341 & 2.0133 & 2.0000 & 0.6668 \\
10s & Huber & -0.4611 & -0.5147 & 0.3728 & 2.0200 & 2.0000 & 0.7092 \\
10s & RankNet & -0.4560 & -0.5144 & 0.3924 & 1.9733 & 2.0000 & 0.6813 \\
10s & MSE+RankNet & -0.4542 & -0.5145 & 0.3575 & 1.9900 & 2.0000 & 0.6805 \\
10s & Huber+ListNet & -0.4539 & -0.5153 & 0.3938 & 1.9600 & 2.0000 & 0.7024 \\
10s & MSE+LambdaRank & -0.4476 & -0.5134 & 0.3835 & 2.0450 & 2.0000 & 0.7074 \\
10s & Huber+LambdaRank & -0.4456 & -0.5136 & 0.3874 & 2.0067 & 2.0000 & 0.7118 \\

\midrule
\multicolumn{8}{l}{\textbf{60s budget}} \\
\midrule
60s & \textbf{VBS} & -0.2807 & 0.0000 & 0.4048 & 2.6367 & 3.0000 & 0.4921 \\
60s & \textbf{SBS}$_{\text{cost | rank}}$ (\textbf{MAOS}) & -0.2048 & 0.0000 & 0.5592 & 2.7400 & 3.0000 & 0.6140 \\
\midrule
60s & EAX & -0.2029 & 0.0000 & 0.3894 & 2.7533 & 3.0000 & 0.4916 \\
60s & LKH & -0.1603 & 0.0000 & 0.4458 & 2.8050 & 3.0000 & 0.5135 \\
60s & CLK & 0.1640 & 0.0000 & 0.7805 & 3.2567 & 3.0000 & 0.8194 \\
60s & Concorde & 0.4227 & 0.0000 & 0.9240 & 3.4450 & 3.0000 & 0.9518 \\

\midrule
60s & \cellcolor{gray!20} \textbf{Huber+ListNet} & \cellcolor{gray!20} \textbf{-0.2548} & 0.0000 & 0.4054 & 2.6900 & 3.0000 & 0.4947 \\
60s & MSE+LambdaRank & -0.2482 & 0.0000 & 0.4424 & 2.7050 & 3.0000 & 0.5336 \\
60s & ListNet & -0.2472 & 0.0000 & 0.4545 & 2.6967 & 3.0000 & 0.5297 \\
60s & MSE+ListNet & -0.2466 & 0.0000 & 0.4542 & 2.7100 & 3.0000 & 0.5371 \\
60s & MSE & -0.2447 & 0.0000 & 0.4255 & \cellcolor{gray!20} \textbf{2.6883} & 3.0000 & 0.5062 \\
60s & MAE+RankNet & -0.2405 & 0.0000 & 0.4611 & 2.6950 & 3.0000 & 0.5232 \\
60s & Huber+RankNet & -0.2213 & 0.0000 & 0.4413 & 2.7200 & 3.0000 & 0.5204 \\
60s & MSE+RankNet & -0.2195 & 0.0000 & 0.4681 & 2.7217 & 3.0000 & 0.5425 \\
60s & Huber+LambdaRank & -0.2154 & 0.0000 & 0.5113 & 2.7183 & 3.0000 & 0.5767 \\
60s & Huber & -0.2131 & 0.0000 & 0.4965 & 2.7450 & 3.0000 & 0.5805 \\
60s & RankNet & -0.2130 & 0.0000 & 0.4593 & 2.7300 & 3.0000 & 0.5352 \\
60s & MAE+ListNet & -0.2048 & 0.0000 & 0.5592 & 2.7400 & 3.0000 & 0.6140 \\
60s & MAE & -0.1944 & 0.0000 & 0.4090 & 2.7683 & 3.0000 & 0.4963 \\
60s & MAE+LambdaRank & -0.1603 & 0.0000 & 0.4458 & 2.8050 & 3.0000 & 0.5135 \\
\bottomrule
\end{tabular}
}
\begin{flushleft}
\scriptsize
\textit{Note.} For the 60s budget, median normalized costs collapse to zero because more than half of the test instances are saturated: 154 of 300 instances have normalized cost zero for all five solvers, and SBS attains the VBS cost on most of instances. 
\end{flushleft}

\end{table}

Table~\ref{tab:overall_perf} summarizes the aggregate performance of the Virtual Best Solver (VBS), the Single Best Solver (SBS), the 5 standalone TSP algorithms, and the GNNAS-TSP models under the 10s and 60s runtime budgets. 
We use mean normalized cost as the primary evaluation criterion since the main objective of the TSP solver portfolio is to minimize final solution cost. 
Mean rank is reported as a complementary metric that reflects the relative position and stability of the selected solver across instances.

Under the 10s budget, the VBS achieves a mean cost of $-0.5069$, while the cost-based SBS is CLK with a mean cost of $-0.4481$. 
Among the AS models, MAE+LambdaRank delivers the lowest mean cost of $-0.4912$, improving upon the SBS by an absolute margin of $0.0431$. 
Its mean rank is $1.9483$, which is highly competitive. 
Although ListNet obtains the lowest mean rank of $1.9350$ among the AS models, its mean cost is worse than MAE+LambdaRank, indicating that optimizing purely for rank does not necessarily yield the best cost-oriented trade-off under a short time budget. 
The 10s standalone-solver results also show a divergence between cost-based and rank-based baselines: CLK is the SBS by mean cost, but MAOS has the best mean rank of $1.9450$ among the candidate solvers. 
Notably, while the best AS models improve the mean cost, their standard deviations in cost are generally higher than that of the SBS.
This higher variance indicates that the selector trades the consistent performance of a single baseline for a more volatile strategy, successfully identifying the best solver on some instances while making sub-optimal yet still strong algorithm choices on others.

Under the 60s budget, the VBS mean cost is $-0.2807$, and the SBS becomes MAOS with a mean cost of $-0.2048$ and a mean rank of $2.7400$. 
EAX's performance is very close to MAOS, with a mean cost of $-0.2029$, indicating that the strongest standalone solvers converge toward similar solution qualities given the extended runtime. 
This convergence leads to a saturation effect where more than half of the test instances yield a normalized cost of zero for all solvers. 
As a result, the median cost is exactly 0.0000 for every method, rendering the median metric identical across all models and useless for distinguishing between them. 
In this regime, the mean cost becomes the sole effective metric for evaluating performance differences. 
Among the learned selectors, Huber+ListNet provides the lowest mean cost of $-0.2548$, improving upon the 60s SBS by $0.0500$. 
Although MSE has the best mean rank of $2.6883$ among selectors, the rank difference from Huber+ListNet at $2.69$ is negligible. 
Since Huber+ListNet maintains a clear cost advantage, it serves as the best AS model under the 60s budget. 
Furthermore, the best 60s selector achieves a cost standard deviation of $0.4054$, which is notably lower than the $0.5592$ standard deviation of the SBS. This indicates that under the extended budget, the learned model not only improves the average performance but also provides greater stability by consistently avoiding the high-cost outliers inherent to the single best baseline.

\subsection{VBS Gap Analysis}
\label{subsec:vbs}

We then evaluate the selector relative to the Virtual Best Solver (VBS) and the Single Best Solver (SBS). 
The analysis is conducted for two lower-is-better metrics: normalized cost and average tied rank. 
Let $M_{\mathrm{AS},i}$ denote the metric value achieved by the Algorithm Selection (AS) models on instance $i$, where $M$ is either normalized cost $C$ or average tied rank $R$. 
For each metric, the SBS is chosen separately as the standalone solver with the best mean value under that metric. 
Thus, the cost-based SBS and rank-based SBS need not be the same solver.

The absolute VBS gap measures the remaining distance from the oracle:
\[
\Delta_i^{\mathrm{VBS}}
=
M_{\mathrm{AS},i}-M_{\mathrm{VBS},i},
\]
where lower values are better and $\Delta_i^{\mathrm{VBS}}=0$ means that AS matches VBS on instance $i$.

We also compute the per-instance VBS gap-closed ratio:
\[
\rho_i
=
\frac{M_{\mathrm{SBS},i}-M_{\mathrm{AS},i}}
     {M_{\mathrm{SBS},i}-M_{\mathrm{VBS},i}}.
\]
This ratio measures the fraction of the available improvement from SBS to VBS that is recovered by AS on instance $i$. 
Under this definition, $\rho_i=1$ means that AS matches VBS, $\rho_i=0$ means that AS matches SBS, and $0<\rho_i<1$ means that AS improves over SBS but does not fully reach VBS. 
A negative value indicates that AS performs worse than SBS.

The per-instance ratio is undefined when SBS and VBS are identical or nearly identical. 
We therefore compute $\rho_i$ only on the non-tie subset:
\[
|M_{\mathrm{SBS},i}-M_{\mathrm{VBS},i}|>\tau,
\quad \tau=10^{-6}.
\]
The tie rate is reported separately as a dataset characteristic. 
A high tie rate means that the fixed SBS already matches VBS on many instances, leaving limited room for any selector to demonstrate improvement. 
For this reason, we report the mean per-instance gap-closed ratio only on the non-tie subset.

In addition, we report an aggregate gap-closed ratio:
\[
\rho_{\mathrm{agg}}
=
\frac{\overline{M}_{\mathrm{SBS}}-\overline{M}_{\mathrm{AS}}}
     {\overline{M}_{\mathrm{SBS}}-\overline{M}_{\mathrm{VBS}}},
\]
where the bars denote means over all test instances. 
Unlike the mean of $\rho_i$, which averages ratios only over the non-tie subset, $\rho_{\mathrm{agg}}$ is computed from the overall mean performance values. 
It therefore summarizes how much of the mean SBS-to-VBS gap is closed by AS across the full test set.

Table~\ref{tab:vbs_gap} reports the VBS gap and gap-closed ratio results for the 10s and 60s budgets. 
In contrast to selecting a separate best learned selector for each metric, we evaluate the same selected overall selector under both normalized cost and average tied rank. 
This is consistent with the model-selection procedure used in the preceding analysis, where the selector is chosen by jointly considering cost and rank rather than by optimizing each metric independently.

For the 10s budget, the selected overall selector is MAE+LambdaRank. 
Under the cost metric, it achieves a mean VBS gap of $0.0157$ and an aggregate gap-closed ratio of $0.7328$, meaning that it closes about $73.3\%$ of the mean SBS-to-VBS cost gap. 
On the non-tie subset, its mean per-instance gap-closed ratio is $0.4873$, and it exactly matches VBS on $83.7\%$ of non-tie instances.

Under the 10s rank metric, the same MAE+LambdaRank selector is compared with the rank-based SBS, MAOS. 
Here, the rank-based tie rate is very high ($90.3\%$), leaving only small numbers of non-tie instances. 
The selector has a mean rank VBS gap of $0.1867$ and a mean per-instance gap-closed ratio of $0.4397$ on the non-tie subset. 
However, its aggregate gap-closed ratio is slightly negative ($-0.0182$), indicating that, in terms of mean rank, it does not improve over the rank-based SBS. 
This is consistent with the Wilcoxon result: MAE+LambdaRank is cost-best and rank-competitive, but it is not significantly better than MAOS in rank.

For the 60s budget, the selected overall selector is Huber+ListNet. 
Under the cost metric, it achieves a mean VBS gap of $0.0260$ and an aggregate gap-closed ratio of $0.6582$. 
The tie rate is high ($96.3\%$), so only few instances have a nonzero SBS--VBS denominator. 
Within this non-tie subset, Huber+ListNet matches VBS on $72.7\%$ of instances.

Under the 60s rank metric, Huber+ListNet achieves a mean rank VBS gap of $0.0533$ and an aggregate gap-closed ratio of $0.4839$. 
Its mean per-instance gap-closed ratio is $0.7576$, again computed over the small non-tie subset. 
Thus, under the 60s budget, the selected selector improves over SBS in both normalized cost and rank, although the high tie rate means that the per-instance ratio statistics should be interpreted cautiously.

Overall, the VBS-gap analysis shows that the selected learned selectors close a substantial fraction of the SBS-to-VBS gap in normalized cost. 
The rank-based analysis is useful as a secondary robustness check, but it is more sensitive to ties and to the choice of rank-based SBS. 
A strong selector should have a small absolute VBS gap, a large positive aggregate gap-closed ratio, a high probability of $\rho_i=1$, and a low probability of $\rho_i<0$.

\begin{table}[!htbp]
\centering
\small
\setlength{\tabcolsep}{4.5pt}
\renewcommand{\arraystretch}{1.15}
\caption{VBS gap and gap-closed ratio summary for 10s and 60s budgets. Columns ``Budg.'', ``Met.'', and ``Loss Fn.'' abbreviate Budget, Performance Metric, and Loss Function. LR and LN refers to LambdaRank and ListNet. The metric $M$ is either normalized cost or average tied rank. The absolute VBS gap is $M_{\mathrm{AS}}-M_{\mathrm{VBS}}$ (lower is better). ``Tie'' is the tie rate, ``Gap'' is the mean VBS gap, $\bar{\rho}$ is the mean gap-closed ratio, and ``Agg. $\rho$'' is the aggregated ratio over all instances. The final column, ``$\Pr$ vector'', reports the four per-instance gap-closed ratio probabilities in this exact order: $\Pr(\rho=1)$, $\Pr(0<\rho<1)$, $\Pr(\rho=0)$, and $\Pr(\rho<0)$.}
\label{tab:vbs_gap}
\begin{tabular}{@{}l | ll |l|rrrr | l@{}}
\toprule
Budg. & Met. & Loss Fn. & SBS & Tie & Gap & $\bar{\rho}$ & Agg. $\rho$ & $\Pr$ vector \\
\midrule
10s & Cost & MAE+LR & CLK & 0.4467 & 0.0157 & 0.4873 & 0.7328 & 0.837/0.090/0.024/0.048 \\
10s & Rank & MAE+LR & MAOS & 0.9033 & 0.1867 & 0.4397 & -0.0182 & 0.414/0.103/0.448/0.035 \\
60s & Cost & Huber+LN & MAOS & 0.9633 & 0.0260 & 0.8181 & 0.6582 & 0.727/0.091/0.182/0.000 \\
60s & Rank & Huber+LN & MAOS & 0.9633 & 0.0533 & 0.7576 & 0.4839 & 0.727/0.091/0.182/0.000 \\
\bottomrule
\end{tabular}
\end{table}

\subsection{Statistical Significance Testing}
\label{subsec:stats}

The loss configurations analysed in this section were selected using the aggregate results on the held-out 300-instance test set reported in Table~\ref{tab:overall_perf}. The paired Wilcoxon analyses below use the same test instances and therefore constitute post-selection exploratory analyses rather than independent confirmatory tests. They quantify paired differences between the selected learned selector and the corresponding Single Best Solver (SBS) within this test sample.

The comparison is performed at the instance level, since each TSP instance has both the performance of the selected learned selector and the corresponding SBS.

For each time budget, we compare the best AS model against SBS using two metrics: normalized true selected cost and selected rank. 
For the cost-based comparison, the paired difference for instance $i$ is defined as
\[
d_i^{\mathrm{cost}}
=
\widetilde{C}_{AS,i}
-
\widetilde{C}_{SBS,i},
\]
where $\widetilde{C}_{AS,i}$ is the normalized true cost obtained by the algorithm selected by the learned selector, and $\widetilde{C}_{SBS,i}$ is the normalized true cost of SBS. For the rank-based comparison, the paired difference is
\[
d_i^{\mathrm{rank}}
=
R_{AS,i}
-
R_{SBS,i}.
\]
Since lower cost and lower rank are better, our hypothesis is directional: the learned selector is better than SBS. We therefore use a one-sided Wilcoxon signed-rank test with
\[
H_0:\operatorname{median}(d_i)=0,
\qquad
H_1:\operatorname{median}(d_i)<0.
\]
Thus, a small one-sided $p$-value indicates that, within this test sample, the selected learned selector tends to achieve lower cost or rank than the corresponding SBS. Because the selector was chosen using the same test-set summaries, these $p$-values are interpreted descriptively and not as independent confirmatory inference.

Table~\ref{tab:wilcoxon_best_selector_vs_sbs} reports the results. Under the 10s budget, the overall best selector, MAE+LambdaRank, significantly outperforms SBS on normalized cost. It improves over the cost-based SBS CLK on 154 instances, ties on 131 instances, and is worse on 15 instances, giving a one-sided Wilcoxon $p$-value of $2.58\times 10^{-20}$. For rank, the same selector is compared with the rank-based SBS MAOS. This comparison is not significant ($p=0.5358$), which is consistent with the aggregate means: MAE+LambdaRank has a slightly higher mean rank than MAOS. Thus, the selected 10s model shows a significant advantage in normalized cost, but not in rank.

Under the 60s budget, the selected overall best selector is Huber+ListNet. It improves over SBS on 9 instances, ties on 287 instances, and is worse on 4 instances. Although most instances are ties, the one-sided Wilcoxon test indicates significant improvement under the directional hypothesis for both normalized cost ($p=0.0374$) and rank ($p=0.0335$). Therefore, the Wilcoxon results provide strong evidence for cost improvement under 10s, and significant cost and rank improvement under 60s.

\begin{table}[htbp]
\centering
\small
\setlength{\tabcolsep}{3pt}
\renewcommand{\arraystretch}{1.1}
\caption{Wilcoxon signed-rank test comparing the best learned selector with SBS. For cost, the test uses normalized true selected cost; for rank, the test uses selected rank. Columns: $N$ is total instances; $N_{\neq0}$ is non-zero difference pairs; $\mu_{\mathrm{Sel}}$ and $\mu_{\mathrm{SBS}}$ are the means of the learned selector and SBS, respectively; $\Delta = \mu_{\mathrm{Sel}} - \mu_{\mathrm{SBS}}$; the columns $+$, $=$, and $-$ report the number of instances where the learned selector is Better, Equal, or Worse than SBS, respectively. The one-sided $p$-value tests the alternative that the learned selector has lower cost or rank; $*$ denotes significance at the 0.05 level.}
\label{tab:wilcoxon_best_selector_vs_sbs}
\begin{tabular}{@{} l | ll | l | rr | rrr | rrr | r @{}}
\toprule
Budg. & Met. & Loss Fn. & SBS & $N$ & $N_{\neq0}$ & $\mu_{\mathrm{Sel}}$ & $\mu_{\mathrm{SBS}}$ & $\Delta$ & $+$ & $=$ & $-$ & $p$ \\
\midrule
10s & Cost & MAE+LR & CLK & 300 & 169 & -0.4912 & -0.4481 & -0.0430 & 154 & 131 & 15 & \cellcolor{gray!20} $<0.001*$ \\
10s & Rank & MAE+LR & MAOS & 300 & 33 & 1.9483 & 1.9450 & 0.0033 & 15 & 267 & 18 & 0.5358 \\
60s & Cost & Huber+LN & MAOS & 300 & 13 & -0.2548 & -0.2048 & -0.0500 & 9 & 287 & 4 & \cellcolor{gray!20} 0.0374* \\
60s & Rank & Huber+LN & MAOS & 300 & 13 & 2.6900 & 2.7400 & -0.0500 & 9 & 287 & 4 & \cellcolor{gray!20} 0.0335* \\
\bottomrule
\end{tabular}
\end{table}

Overall, the one-sided Wilcoxon signed-rank tests provide partial support for the hypothesis that the selected learned selector improves over SBS. Under the 10s budget, the selected selector shows a highly significant improvement in normalized cost, but it does not significantly improve over the rank-based SBS in selected rank. This is consistent with the aggregate results: the selected 10s model achieves the best mean cost while remaining competitive, but not best, in rank. Under the 60s budget, the selected selector achieves significant improvement over SBS for both normalized cost and rank, although the effect is based on a small number of non-tied instance pairs because most instances are ties.

\subsection{Algorithm Selection Behavior}
\label{subsec:behavior}




To better understand the behavior of the AS models, we analyze how frequently each method assigns test instances to the five candidate solvers: CLK, EAX, LKH, MAOS, and Concorde. This analysis is especially important because many solvers obtain identical final costs on some instances, particularly under the 60s budget. Therefore, a naive VBS selection frequency based on a single argmin can be misleading: if multiple solvers are tied for the best observed cost, assigning the whole instance to one solver may overstate that solver's importance.

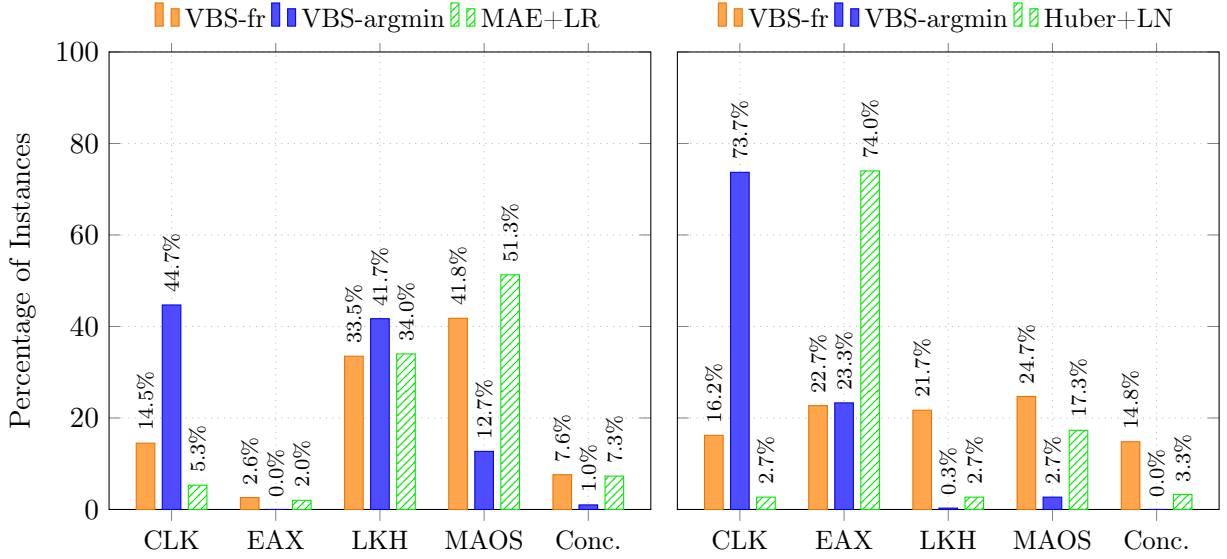
\begin{figure}[htbp]
\centering
\begin{tikzpicture}

\begin{axis}[
    name=left,
    width=0.45\textwidth,
    height=0.38\textwidth,
    scale only axis,
    ybar=2.8pt,
    bar width=7pt,
    ymin=0, ymax=100,
    enlarge x limits=0.15,
    ylabel={Percentage of Instances},
    symbolic x coords={CLK,EAX,LKH,MAOS,Conc.},
    xtick=data,
    xticklabel style={font=\small},
    ytick={0,20,40,60,80,100},
    yticklabel style={/pgf/number format/fixed},
    grid=major,
    grid style={dotted,gray!50},
    axis line style={-},
    title={10s Time Budget},
    title style={yshift=-1mm},
    legend style={
        at={(0.5,1.02)},
        anchor=south,
        legend columns=3,
        draw=none,
        font=\small
    },
    nodes near coords,
    nodes near coords align={vertical},
    every node near coord/.append style={
        font=\scriptsize,            
        rotate=90,
        anchor=west,
        xshift=1pt
    }
]

\addplot[
    fill=orange!70,
    draw=orange!90!black,
    point meta=explicit symbolic
] coordinates {
    (CLK,14.5) [14.5\%]
    (EAX,2.6) [2.6\%]
    (LKH,33.5) [33.5\%]
    (MAOS,41.8) [41.8\%]
    (Conc.,7.6) [7.6\%]
};

\addplot[
    fill=blue!70,
    draw=blue!90!black,
    point meta=explicit symbolic
] coordinates {
    (CLK,44.7) [44.7\%]
    (EAX,0.0) [0.0\%]
    (LKH,41.7) [41.7\%]
    (MAOS,12.7) [12.7\%]
    (Conc.,1.0) [1.0\%]
};

\addplot[
    fill=green!70,
    draw=green!90!black,
    pattern=north east lines,
    pattern color=green!90!black,
    point meta=explicit symbolic
] coordinates {
    (CLK,5.3) [5.3\%]
    (EAX,2.0) [2.0\%]
    (LKH,34.0) [34.0\%]
    (MAOS,51.3) [51.3\%]
    (Conc.,7.3) [7.3\%]
};

\legend{VBS-fr, VBS-argmin, MAE+LR}
\end{axis}

\begin{axis}[
    name=right,
    at={(left.south east)},
    anchor=south west,
    xshift=10pt,
    width=0.45\textwidth,
    height=0.38\textwidth,
    scale only axis,
    ybar=2.8pt,
    bar width=7pt,
    ymin=0, ymax=100,
    enlarge x limits=0.15,
    symbolic x coords={CLK,EAX,LKH,MAOS,Conc.},
    xtick=data,
    xticklabel style={font=\small},
    ytick={0,20,40,60,80,100},
    yticklabels={},
    grid=major,
    grid style={dotted,gray!50},
    axis line style={-},
    title={60s Time Budget},
    title style={yshift=-1mm},
    legend style={
        at={(0.5,1.02)},
        anchor=south,
        legend columns=3,
        draw=none,
        font=\small
    },
    nodes near coords,
    nodes near coords align={vertical},
    every node near coord/.append style={
        font=\scriptsize,
        rotate=90,
        anchor=west,
        xshift=1pt
    }
]

\addplot[
    fill=orange!70,
    draw=orange!90!black,
    point meta=explicit symbolic
] coordinates {
    (CLK,16.2) [16.2\%]
    (EAX,22.7) [22.7\%]
    (LKH,21.7) [21.7\%]
    (MAOS,24.7) [24.7\%]
    (Conc.,14.8) [14.8\%]
};

\addplot[
    fill=blue!70,
    draw=blue!90!black,
    point meta=explicit symbolic
] coordinates {
    (CLK,73.7) [73.7\%]
    (EAX,23.3) [23.3\%]
    (LKH,0.3) [0.3\%]
    (MAOS,2.7) [2.7\%]
    (Conc.,0.0) [0.0\%]
};

\addplot[
    fill=green!70,
    draw=green!90!black,
    pattern=north east lines,
    pattern color=green!90!black,
    point meta=explicit symbolic
] coordinates {
    (CLK,2.7) [2.7\%]
    (EAX,74.0) [74.0\%]
    (LKH,2.7) [2.7\%]
    (MAOS,17.3) [17.3\%]
    (Conc.,3.3) [3.3\%]
};

\legend{VBS-fr, VBS-argmin, Huber+LN}
\end{axis}

\end{tikzpicture}
\caption{Solver selection frequencies as percentages over 300 test instances for the 10‑second (left) and 60‑second (right) time budgets. Bars show the percentage of instances assigned to each solver by VBS‑fractional (VBS-fr), VBS‑argmin, and the corresponding AS model. VBS-argmin assigns each instance to one best solver while VBS-fractional distributes credit equally among all solvers tied for the best observed cost.}
\label{fig:solver_selection_frequency}
\end{figure}

To address this issue, Figure~\ref{fig:solver_selection_frequency} reports both VBS-argmin and VBS-fractional frequencies. 
VBS-argmin assigns each instance to one solver with the lowest observed cost, while VBS-fractional distributes the credit equally among all solvers tied for the best observed cost. The VBS-fractional frequency is more consistent with our average tied-rank evaluation metric, because it avoids artificially separating solvers that achieve the same final solution quality. The selection frequencies are computed from the solver selected by the model, where the selected solver is the one with the lowest predicted cost.

Under the 10s budget, the VBS-argmin distribution appears concentrated on CLK and LKH. However, the VBS-fractional distribution shows a different pattern: MAOS receives the largest fraction of tied-best credit, followed by LKH. This indicates that multiple solvers often share the best observed cost, and a single argmin label does not fully represent solver behavior. The selected learned selector, MAE+LambdaRank, assigns most instances to MAOS and LKH, which is broadly consistent with the VBS-fractional distribution. In contrast, the cost-based SBS always selects CLK, and the rank-based SBS always selects MAOS. Therefore, the learned selector does not collapse to a single fixed solver; instead, it uses different solvers across instances and departs from the static SBS choices.

Under the 60s budget, the difference between VBS-argmin and VBS-fractional becomes even more important. VBS-argmin assigns 73.7\% of instances to CLK, which may suggest that CLK dominates at 60s. However, the VBS-fractional distribution is much more balanced across the five solvers. This means that many 60s instances contain several solvers tied for the best final cost. The selected learned selector, Huber+ListNet, chooses EAX on 74.0\% of the test instances, which is very different from the VBS-argmin distribution. This mismatch should not be interpreted as a failure of the selector, because VBS-argmin ignores tied-best solvers. Instead, it shows that the learned selector often chooses a solver different from the single VBS-argmin label while still achieving a tied-best or near-best final cost.

To further clarify this point, Table~\ref{tab:tie_subset_diagnostics} separates test instances into two subsets. The subset SBS=VBS contains instances where the cost-based SBS already reaches the VBS cost, so algorithm selection has no cost gap to close. The subset SBS$>$VBS contains instances where the VBS improves over SBS, so algorithm selection has meaningful room for improvement.

\begin{table}[!htbp]
\centering
\caption{Performance of the best learned selector on tie and non-tie subsets. The subset SBS=VBS contains instances where the cost-based SBS already matches the VBS cost, while SBS$>$VBS contains instances where algorithm selection has room to improve over SBS. Cost values are normalized costs.}
\label{tab:tie_subset_diagnostics}
\resizebox{\textwidth}{!}{
\begin{tabular}{llrrrrrr}
\toprule
Budget & Subset & \# Inst. & Selector Cost & SBS Cost & VBS Cost & Tied-Best Rate & Gap Closed \\
\midrule
10s & SBS=VBS & 134 & -0.385 & -0.410 & -0.410 & 94.8\% & -- \\
10s & SBS$>$VBS & 166 & -0.577 & -0.479 & -0.585 & 83.7\% & 48.7\% \\
\midrule
60s & SBS=VBS & 289 & -0.254 & -0.272 & -0.272 & 98.6\% & -- \\
60s & SBS$>$VBS & 11 & -0.282 & 1.558 & -0.513 & 72.7\% & 81.8\% \\
\bottomrule
\end{tabular}
}
\end{table}

Table~\ref{tab:tie_subset_diagnostics} shows that under the 10s budget, 166 out of 300 test instances have an SBS--VBS gap. On this informative subset, MAE+LambdaRank closes 48.7\% of the cost gap on average and selects a tied-best solver on 83.7\% of instances. This supports the conclusion that the learned selector can exploit meaningful solver-performance differences under the shorter budget.

Under the 60s budget, only 11 out of 300 test instances have an SBS--VBS gap, confirming that solver outcomes become highly tied at the longer budget. Nevertheless, on this small but informative subset, Huber+ListNet closes 81.8\% of the cost gap and selects a tied-best solver on 72.7\% of instances. On the remaining 289 SBS=VBS instances, the selected solver is tied-best on 98.6\% of instances. Therefore, although the overall opportunity for algorithm selection is limited at 60s because most solvers converge to the same cost, the learned selector still performs better than SBS on the instances where meaningful solver-performance differences remain.

Overall, the selection-frequency analysis shows that the learned selectors do not merely reproduce SBS behavior. Under 10s, the selected model mainly selects MAOS and LKH rather than always choosing the cost-based SBS, CLK. Under 60s, the selected model frequently chooses EAX instead of the SBS solver, MAOS, but this behavior is reasonable once tied-best solutions are considered. The key conclusion is that the learned selectors exploit instance-dependent solver preferences while operating in a setting where many solvers can obtain identical final costs, especially under the longer runtime budget.

\subsection{Low Dimensional Performance Structure}
\label{subsec:structure}




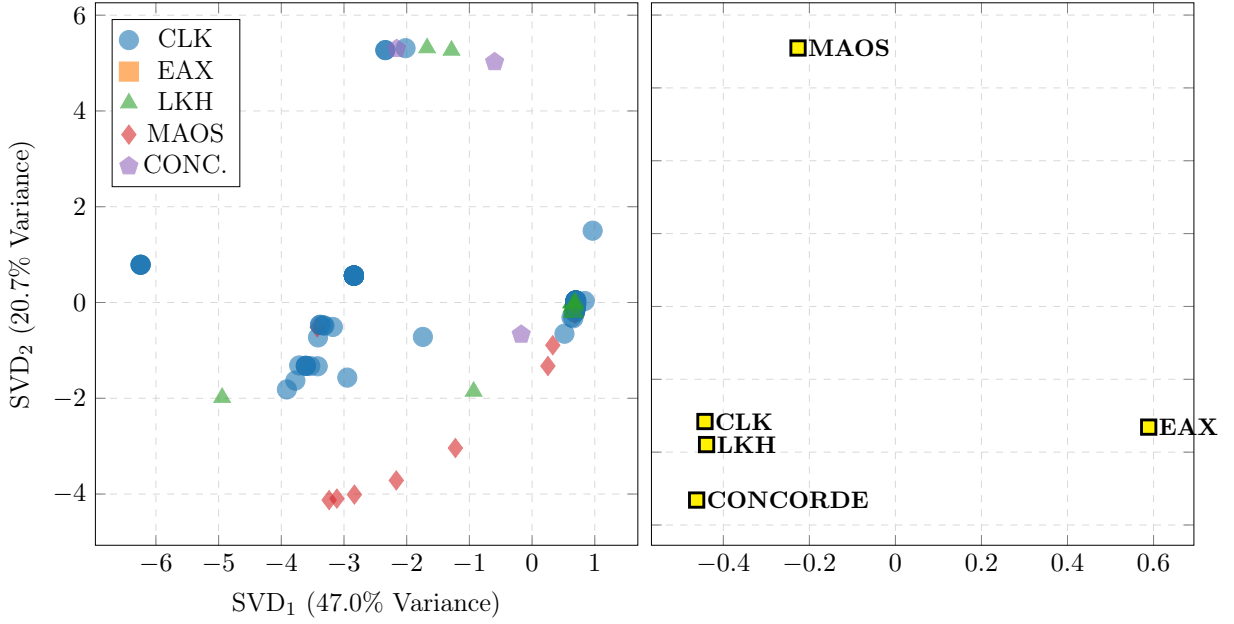
\begin{figure}[htb]
    \centering
    
\begin{tikzpicture}[
    scale=0.9, transform shape]
\begin{groupplot}[
    legend pos=north west, 
    group style={
        group size=2 by 1,
        horizontal sep=0.2cm,
        y descriptions at=edge left,
    },
    width=0.6\textwidth,
    height=0.6\textwidth,
    grid=major,
    grid style={dashed, gray!30},
]

\nextgroupplot[
    xlabel={SVD$_1$ (47.0\% Variance)},
    ylabel={SVD$_2$ (20.7\% Variance)},
    scatter/classes={
        0={mark=*, color=colorCLK},
        1={mark=square*, color=colorEAX},
        2={mark=triangle*, color=colorLKH},
        3={mark=diamond*, color=colorMAOS},
        4={mark=pentagon*, color=colorCONCORDE}
    },
]

\addplot[
    scatter,
    only marks,
    scatter src=explicit symbolic,
    mark size=4pt,
    opacity=0.6,
]
table[x=x, y=y, meta=best_alg_id] {GNNAS-TSP-TestInst-5Algs-ASmethods-PerformanceData-10s-COST-instances-svd.dat};

\addlegendentry{CLK}
\addlegendentry{EAX}
\addlegendentry{LKH}
\addlegendentry{MAOS}
\addlegendentry{CONC.}

\nextgroupplot[
]

\addplot[only marks, mark=square*, mark size=3pt,
         mark options={fill=yellow, draw=black, very thick},
         nodes near coords={\pgfplotspointmeta},
         point meta=explicit symbolic,
         every node near coord/.append style={
             font=\bfseries\small,
             anchor=west,
             text opacity=1,
         }]
    table[x=x, y=y, meta=alg_name] {GNNAS-TSP-TestInst-5Algs-ASmethods-PerformanceData-10s-COST-algorithms-svd.dat};

\end{groupplot}
\end{tikzpicture}

    \caption{SVD visualization of the 10s normalized-cost matrix. Left: instances colored by hard VBS solver. Right: algorithm loadings.}
\label{fig:svd_10s}
\end{figure}

\begin{figure}[htb]
    \centering

\begin{tikzpicture}[
    scale=0.9, transform shape]
\begin{groupplot}[
    legend pos=south west, 
    group style={
        group size=2 by 1,
        horizontal sep=0.2cm,
        y descriptions at=edge left,
    },
    width=0.6\textwidth,
    height=0.6\textwidth,
    grid=major,
    grid style={dashed, gray!30},
]

\nextgroupplot[
    xlabel={SVD$_1$ (39.1\% Variance)},
    ylabel={SVD$_2$ (27.2\% Variance)},
    scatter/classes={
        0={mark=*, color=colorCLK},
        1={mark=square*, color=colorEAX},
        2={mark=triangle*, color=colorLKH},
        3={mark=diamond*, color=colorMAOS},
        4={mark=pentagon*, color=colorCONCORDE}
    },
]

\addplot[
    scatter,
    only marks,
    scatter src=explicit symbolic,
    mark size=4pt,
    opacity=0.6,
]
table[x=x, y=y, meta=best_alg_id] {GNNAS-TSP-TestInst-5Algs-ASmethods-PerformanceData-60s-COST-instances-svd.dat};

\addlegendentry{CLK}
\addlegendentry{EAX}
\addlegendentry{LKH}
\addlegendentry{MAOS}
\addlegendentry{CONC.}


\nextgroupplot[
    clip=false,   
]

\addplot[
    only marks,
    mark=square*, mark size=3pt,
    mark options={fill=yellow, draw=black, very thick},
    nodes near coords={\pgfplotspointmeta},
    point meta=explicit symbolic,
    visualization depends on={value \coordindex \as \idx},
    every node near coord/.append style={
        font=\bfseries\small,
        fill=white,
        anchor=\ifcase\idx east\or north\or south\or west\or east\fi,
        xshift=\ifcase\idx {-0.2cm}\or {0cm}\or {1cm}\or {0.4cm}\or {-0.3cm}\fi,
        xshift=\ifcase\idx {0cm}\or {-1.8cm}\or {0.5cm}\or {0.5cm}\or {0cm}\fi,
    }
] table[x=x, y=y, meta=alg_name] {GNNAS-TSP-TestInst-5Algs-ASmethods-PerformanceData-60s-COST-algorithms-svd.dat};
\end{groupplot}
\end{tikzpicture}

    \caption{SVD visualization of the 60s normalized-cost matrix. Left: instances colored by hard VBS solver. Right: algorithm loadings.}
\label{fig:svd_60s}
\end{figure}

To further examine the global structure of the algorithms' behavior, we apply Singular Value Decomposition (SVD) to the standardized instance--algorithm normalized-cost matrix. 
Let $\mathbf{Y} \in \mathbb{R}^{N \times m}$ denote this matrix where $N$ is the number of test instances and $m=5$ is the number of candidate solvers. 
The SVD factorizes $\mathbf{Y}$ as $\mathbf{Y} = \mathbf{U} \mathbf{\Sigma} \mathbf{V}^\top$, where $\mathbf{U} \in \mathbb{R}^{N \times m}$ and $\mathbf{V} \in \mathbb{R}^{m \times m}$ are orthogonal matrices containing the left and right singular vectors, respectively, and $\mathbf{\Sigma}$ is a diagonal matrix of singular values. 
$\mathbf{U}$ represents the instances while $\mathbf{V}$ contains latent features for those candidate algorithms.
The first two SVD components are used for visualization.
In the instance plots, each point represents one TSP instance and is colored by its hard VBS solver. 
In the algorithm plots, each point represents one candidate solver. 
Instances that are close to each other have similar solver-performance profiles while solvers that are close to each other have similar performance patterns across the benchmark.

Figure~\ref{fig:svd_10s} shows the 10s SVD visualization. 
In the algorithm-loading plot, CLK and LKH are located close to each other, suggesting that they have similar normalized-cost behavior across instances. 
Concorde lies in the same broad region but is separated along the second SVD component, while MAOS and EAX occupy distinct regions. 
This indicates that the solver portfolio contains complementary performance patterns rather than a single homogeneous group of solvers.

The 10s instance-score plot shows that many instances form a dense central cluster, while several smaller groups and outliers appear away from this region. 
The VBS colors are not perfectly separated by the first two SVD components, indicating that the oracle-best solver cannot be explained by a simple two-dimensional partition. 
Nevertheless, the spread of points suggests that the benchmark contains heterogeneous solver-performance profiles. 
Under the 10s budget, EAX does not appear as a VBS color because it is never the hard VBS solver for any instance, although it is still included in the SVD algorithm embedding.

Figure~\ref{fig:svd_60s} shows the corresponding visualization for the 60s budget. 
In the algorithm-loading plot, CLK and Concorde are strongly separated from the remaining solvers, while EAX and LKH appear close to each other and MAOS lies between these groups. 
This suggests that the longer time budget changes the relative solver-performance geometry and creates a different set of solver contrasts from the 10s case.

The 60s instance-score plot also shows structured heterogeneity. 
Most instances with EAX as the hard VBS solver lie along a visible curved region on the right side of the plot, while several CLK-VBS instances appear in more separated regions. 
As in the 10s case, the VBS labels are not completely separated, but the distribution is far from uniform. 
Concorde does not appear as a VBS color under the 60s budget since it is never the hard VBS solver for any instance. 

Overall, the SVD visualization supports the motivation for AS. 
The candidate solvers occupy distinct regions in the algorithm-loading space, and the instances exhibit heterogeneous performance profiles. 
This indicates that solver performance varies systematically across the benchmark rather than one solver uniformly dominating all instances. 
Therefore, a learned selector has meaningful instance-dependent structure to exploit.

\subsection{Summary of Findings}
\label{subsec:summary}

The experimental results show that the effectiveness of the proposed GNN-based Algorithm Selection (GNNAS) system for the Traveling Salesman Problem (GNNAS-TSP) depends on both the runtime budget and the evaluation criterion. 
Across the two budgets, normalized cost and average tied rank do not always identify the same strongest baseline or selector. Under the 10s budget, CLK is the SBS by mean normalized cost, while MAOS is the strongest fixed solver by mean rank. This distinction is important because normalized cost measures absolute solution quality, whereas average tied rank measures relative stability within the solver portfolio.

Under the 10s budget, MAE+LambdaRank is identified as the best overall loss function to build an AS model because it achieves the lowest mean normalized cost among learned selectors while maintaining competitive rank performance. Its cost improvement over the cost-based SBS CLK is substantial and statistically significant: it improves on 154 instances, ties on 131, and is worse on 15, with a one-sided Wilcoxon $p$-value of $2.58\times 10^{-20}$. However, when evaluated by rank against the rank-based SBS MAOS, the same selector does not achieve a significant improvement ($p=0.5358$). This confirms that the 10s selector is primarily cost-effective rather than rank-dominant.

The VBS-gap analysis gives a more detailed view of this behavior. For the 10s cost metric, MAE+LambdaRank closes about $73.3\%$ of the aggregate SBS-to-VBS gap, indicating that it recovers a large fraction of the possible mean cost improvement over SBS. In contrast, for the 10s rank metric, its aggregate gap-closed ratio is slightly negative ($-0.0182$), meaning that it does not improve over the rank-based SBS in mean rank. This is consistent with the significance test and with the aggregate performance table: the selected 10s model is cost-best and rank-competitive, but not rank-best.

Under the 60s budget, Huber+ListNet significantly outperforms SBS in both normalized cost ($p=0.0374$) and rank ($p=0.0335$)according to the Wilcoxon signed-rank test. This indicates that the learned selector can still exploit residual solver-performance differences even when the time budget is longer. However, the high tie rate shows that many instances no longer provide meaningful room for performing AS since several solvers converge to the same or nearly identical solution quality. Therefore, the 60s result should be interpreted as statistically significant but concentrated on the informative non-tie subset.

The VBS-gap results support the same conclusion. 
Under the 60s budget, Huber+ListNet closes $65.8\%$ of the aggregate SBS-to-VBS gap in normalized cost and $48.4\%$ of the aggregate gap in average tied rank. On the non-tie subset, it often reaches the VBS performance, but the high tie rate indicates that many instances already leave little room for improvement over SBS. Therefore, the 60s results provide evidence of useful learned selection behavior, while also showing that the available improvement over the fixed baseline is concentrated in a small subset of instances.

The solver selection frequency analysis shows that the built AS models do not simply reproduce SBS behavior. 
Under 10s, the VBS-argmin distribution is concentrated on CLK and LKH, but the tie-aware VBS-fractional distribution assigns the largest shares to MAOS and LKH. 
The MAE+LambdaRank based AS follows this tie-aware pattern more closely by selecting MAOS and LKH most often. 
Under 60s, VBS-argmin is heavily concentrated on CLK, but VBS-fractional is much more balanced across solvers, reflecting the high frequency of tied-best solutions. In this setting, Huber+ListNet selects EAX most frequently; this differs from VBS-argmin but remains reasonable because many solvers share the same final cost.

Finally, the SVD-based visualization provides additional evidence that the solver portfolio has structured heterogeneity. 
The algorithm-loading plots show that the five solvers occupy distinct regions, indicating different performance patterns across instances. 
The instance plots show that instances are spread across the latent SVD space rather than forming one uniform group, although VBS labels are not perfectly separated by the first two components. 
This suggests that the benchmark contains meaningful solver-performance structure, but that the structure is not trivially separable in a low-dimensional projection.

Overall, the results suggest that a GNN based AS can improve a strong standalone TSP solver, especially in normalized cost. Under the 10s budget, the main benefit is a strong and statistically significant cost improvement, while rank improvement is not significant. Under the 60s budget, the AS models improve over SBS in both cost and rank, although the high tie rate limits the number of informative paired comparisons. Taken together, the results support the use of GNN-based AS for exploiting instance-dependent solver behavior, while also showing that its benefit depends strongly on the metric, tie structure, and runtime budget.

\section{Conclusion and Future Work}
\label{sec:conclusiont_futurework}

\subsection{Summary}


In this work, we proposed \textsc{GNNAS-TSP}, a Graph Neural Network (GNN) based Algorithm Selection (AS) framework for the Traveling Salesman Problem (TSP). 
Building upon the AS paradigm, our approach learns instance representations directly from raw TSP graphs using GNNs, thereby avoiding manual feature engineering while exploiting the inherent structural properties of the problem.

We evaluated \textsc{GNNAS-TSP} on a portfolio of known TSP algorithms, including Chained Lin--Kernighan (CLK), Edge Assembly Crossover (EAX), Lin--Kernighan--Helsgaun (LKH), and the exact solver Concorde, together with MAOS as a strong portfolio-based baseline. Experiments were conducted under two fixed time budgets of 10 seconds and 60 seconds. Algorithm selection was formulated as a cost prediction and ranking task, and a range of loss functions were investigated, including MSE, RankNet, ListNet, LambdaRank, and their variants.

Across those two budget alternatives, the selected learned selectors improve over static solver choice primarily in normalized cost. 
Under the 10s budget, MAE+LambdaRank achieves a strong and statistically significant cost improvement over the cost-based SBS, although it does not significantly improve over the rank-based SBS. Under the 60s budget, Huber+ListNet significantly improves over SBS in both cost and average tied rank, but the high tie rate indicates that the improvement is concentrated on the small subset of instances where solver-performance differences remain.

Beyond aggregate performance metrics, analyses of solver selection frequencies and low-dimensional embeddings reveal that effective selectors exploit complementary solver strengths rather than concentrating exclusively on the single best solver. Models that better approximate the oracle selection distribution tend to exhibit more robust instance-wise performance. Overall, these findings highlight the effectiveness of graph neural network based algorithm selection as a meta-solving strategy for the TSP, particularly in regimes where solver performance heterogeneity can be reliably identified.

\subsection{Limitations and Future Research Directions}

The present study focuses on AS under fixed time budgets yet budget requirements or resources availabilities may vary, suggesting anytime behavior. 
A natural extension of this work is therefore to move from static, budget-specific AS toward fully anytime AS \citep{huerta2022aaas}. 
One promising direction is the adoption of sequential or temporal Graph Neural Network (GNN) models to capture time-evolving representations in graph-structured data. 
By explicitly modeling solver performance trajectories over time, such models could enable dynamic solver switching and budget-aware decision making during execution~\citep{pareja2020evolvegcn,rossi2020temporal}.

In this work, we adopt the TSP instance collection introduced by~\citet{huerta2022aaas}, which aggregates instances from TSPLIB and several established generators designed to support anytime algorithm selection. While the original dataset provides full anytime performance traces, we repurpose this instance set by extracting solver performance at fixed runtime cutoffs and formulating the problem as instance-wise selection under given time budgets. 
Future work will evaluate the generalization of \textsc{GNNAS-TSP} on additional benchmark collections, especially with larger instances, and alternative distributions, in order to assess robustness across broader problem scales and structural characteristics.

From a modeling perspective, we also plan to explore more expressive architectures, such as GNN--Transformer hybrids and attention-based graph models, which may better capture long-range dependencies and global structural patterns in large TSP instances. In addition, sparse graph constructions such as $k$-nearest-neighbor (kNN) graphs represent potential for improving scalability while preserving essential geometric information.

More broadly, this work opens several directions for future research at the intersection of AS and learning-based optimization, including GNN-based adaptive portfolio design, multi-objective AS, and tighter integration between learning models and classical heuristic solvers.

\section*{CRediT Author Statement}

\textbf{Zhaoxuan Li}: Methodology, Software, Validation, Formal Analysis, Writing – Original Draft, Writing – Review \& Editing.
\textbf{Jiale Yang}: Methodology, 
Software, Validation, Formal Analysis, Writing – Original Draft, Writing – Review \& Editing.
\textbf{Yifei Lu}: Validation, Writing – Original Draft, Writing – Review \& Editing.
\textbf{Mustafa M\i s\i r}: Conceptualization, Supervision, Methodology, Validation, Writing – Original Draft, Writing – Review \& Editing.

\section*{Acknowledgements}
The work has been supported by the Duke Kunshan University Undergraduate Summer Research Scholars Program (SRS).

%
%
\bibliographystyle{apalike-ejor}
\bibliography{references}

\end{document}